\documentclass[letterpaper, 10 pt, conference]{ieeeconf}  

\IEEEoverridecommandlockouts                              
\makeatletter
\let\NAT@parse\undefined
\makeatother
\usepackage[numbers,sort&compress]{natbib}
\usepackage{amsmath,amsfonts,amssymb}
\let\labelindent\relax
\usepackage{enumitem}
\usepackage{bm}
\usepackage{prettyref}
\usepackage{graphicx}
\usepackage{mdframed}
\usepackage{wrapfig}
\usepackage{xspace}
\usepackage{subfloat}
\usepackage{subcaption}
\usepackage{cuted}
\usepackage{array,multirow}
\usepackage{booktabs}
\usepackage{mathtools}
\usepackage{bbm}
\usepackage{leftindex}
\usepackage[dvipsnames,table]{xcolor}
\usepackage{cellspace}
\usepackage[T1]{fontenc}
\usepackage{comment}
\usepackage{url}
\usepackage{makecell}
\usepackage{hhline}
\usepackage{cuted}
\usepackage[
    bookmarks=true,
    colorlinks=true,
    linkcolor=brown,
    citecolor=RoyalPurple!75,
    urlcolor=Maroon
]{hyperref}
\usepackage{algorithmic}
\usepackage{algorithm}
\usepackage{multirow}
\newcommand{\ourshort}{\texttt{\textcolor{Maroon}{\textbf{FDP}}}\xspace}
\newcommand{\ourweb}{\href{https://factorized-diffusion-policy.github.io}{factorized-diffusion-policy.github.io}\xspace}

\begin{document}

\title{\LARGE \bf
Flexible Multitask Learning with
Factorized Diffusion Policy
}

\author{
Chaoqi Liu$^{1}$\quad
Haonan Chen$^{2}$\quad
Sigmund H. Høeg$^{3*}$\quad
Shaoxiong Yao$^{1*}$\\
Yunzhu Li$^{4}$\quad
Kris Hauser$^{1}$\quad
Yilun Du$^{2}$
\thanks{
$^{1}$University of Illinois at Urbana-Champaign
$^{2}$Harvard University
$^{3}$Norwegian University of Science and Technology
$^{4}$Columbia University
Corresponding author: {\tt\small chaoqil2@illinois.edu}.
}
}

\maketitle

\begin{abstract}

Multitask learning poses significant challenges due to the highly multimodal and diverse nature of robot action distributions. However, effectively fitting policies to these complex task distributions is often difficult, and existing monolithic models often underfit the action distribution and lack the flexibility required for efficient adaptation. We introduce a novel modular diffusion policy framework that factorizes complex action distributions into a composition of specialized diffusion models, each  capturing a distinct sub-mode of the behavior space for a more effective overall policy. In addition, this modular structure enables flexible policy adaptation to new tasks by adding or fine-tuning components, which inherently mitigates catastrophic forgetting. Empirically, across both simulation and real-world robotic manipulation settings, we illustrate how our method consistently outperforms strong modular and monolithic baselines. Website: \ourweb.

\end{abstract}

\IEEEpeerreviewmaketitle


\section{Introduction}

Imitation learning has emerged as a powerful paradigm for acquiring complex robotic manipulation skills~\cite{chi2024diffusionpolicy,wang2024poco,høeg2025hybriddiffusionsimultaneoussymbolic,chen2025bimanual, chen2025multimodalmanipulationmultimodalpolicy}. However, extending this success to \textit{multitask} settings remains a significant challenge. As the variety of tasks increases, the underlying action distribution becomes highly multimodal and diverse, often involving distinct control strategies across different objects. Traditional monolithic policies often struggle to generalize across tasks, represent multiple behavior modes, or adapt efficiently to new skills~\cite{wang2024poco, ha_scaling_2023,chang2020sight}.

To address these limitations, modular policy architectures, most notably Mixture-of-Experts (MoE) models~\cite{shazeer_outrageously_2017, lin2024moma}, have emerged as a promising direction. By decomposing the policy into specialized components, modular methods improve scalability and reuse across tasks~\cite{wang2024poco, yang2020multi, wang2024sparse_dp_moe, reuss2024mode, chen2025multimodalmanipulationmultimodalpolicy}. Yet, existing MoE-based approaches often suffer from training instability~\cite{lin2024moma}, lack a principled probabilistic formulation, and produce expert modules with unclear or overlapping roles~\cite{wang2024sparse_dp_moe, du2024reducereuserecyclecompositional}, limiting their interpretability.

\begin{figure}[t]
    \centering
    \includegraphics[width=.98\columnwidth]{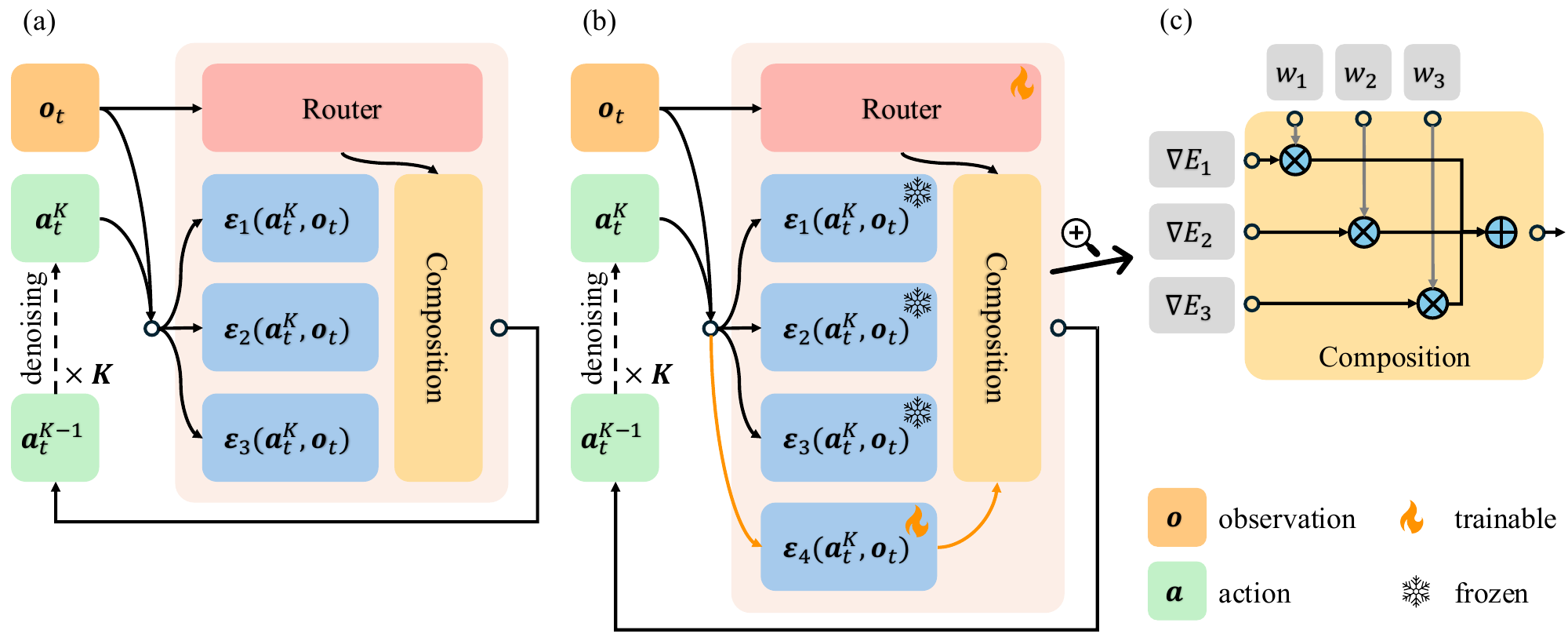}
    \caption{\textbf{Overview of \ourshort}. \textbf{(a)} Given an observation $\mathbf{o}_t$, multiple diffusion experts predict score estimates $\boldsymbol{\varepsilon}_i(\mathbf{a}_t^K, \mathbf{o}_t)$ at each denoising step. A lightweight router network computes observation-dependent weights $\{w_i\}$, which are used to compose the final score as a weighted sum (see \textbf{(c)}). The composed score guides the iterative denoising process over $K$ steps to generate an action $\mathbf{a}_t$. \textbf{(b)} This compositional structure enables \ourshort to model complex multimodal distributions and supports modular adaptation via selective tuning or addition of diffusion components.}
    \label{fig:our_pipeline}
\end{figure}

We propose \textit{Factorized Diffusion Policy} (\ourshort), a simple yet effective modular policy architecture. \ourshort decomposes the policy into multiple diffusion components (Fig.~\ref{fig:our_pipeline}a), each capturing a distinct behavioral mode, which are dynamically composed at inference time via an observation-conditioned router (Fig.~\ref{fig:our_pipeline}c). Instead of discrete expert selection as in standard MoE architectures, \ourshort uses continuous score aggregation, enabling stable training, preventing routing imbalance, and promoting clearer specialization across components. \ourshort is grounded in compositional diffusion modeling~\cite{du2024reducereuserecyclecompositional, du2024compositional, liu2023compositionalvisualgenerationcomposable}, where aggregating scores corresponds to sampling from the product of distributions, providing a principled probabilistic interpretation and a natural formulation as constraint satisfaction. The modular structure further enables efficient task adaptation: we extend the policy by introducing new diffusion components initialized via upcycling~\cite{lin2024moma} from existing components (Fig.~\ref{fig:our_pipeline}b), allowing efficient skill expansion without retraining the entire policy. This factorization improves multitask learning and supports scalable adaptation.

We validate \ourshort through extensive experiments in simulation benchmarks MetaWorld~\cite{yu2021metaworldbenchmark} and RLBench~\cite{james2019rlbenchbenchmark}, and further demonstrate its practical benefits in real-world robotic manipulation. Our contributions are summarized as follows: (1) We introduce a modular diffusion policy architecture that composes specialized components via observation-conditioned compositional sampling. (2) We demonstrate that our compositional framework improves multitask performance and enables sub-skill decomposition across diffusion modules. (3) We propose a simple and effective strategy for adapting to new tasks by selectively tuning or augmenting existing components, achieving superior sample efficiency and modular reuse.


\section{Related Works}

\subsubsection{\textbf{Diffusion Models for Robotics}}
Diffusion models have emerged as a powerful tool for modeling complex distributions, achieving strong performance in image~\cite{ho2020ddpm, pmlr-v139-nichol21a, ramesh2022hierarchicaltextconditionalimagegeneration} and video generation~\cite{ho_video_2022, wang2025learningrealworldactionvideodynamics}. Their stable training and generative flexibility have led to increasing adoption in robotic domains, including video-conditioned policy learning~\cite{du_learning_2023, ajay_compositional}, grasp synthesis~\cite{urainSEDiffusionFieldsLearning2023a}, bimanual manipulation~\cite{chen2025bimanual}, tool use~\cite{chen2025toolasinterface}, trajectory planning~\cite{jannerPlanningDiffusionFlexible2022a, ajayConditionalGenerativeModeling2023, carvalhoMotionPlanningDiffusion2023}, and closed-loop visuomotor control. Diffusion Policy (DP)~\cite{chi2024diffusionpolicy} demonstrated that diffusion models can be used to learn reactive visuomotor policies from demonstrations, achieving state-of-the-art performance in single-task imitation learning.

\subsubsection{\textbf{Multitask Imitation Learning and Adaptation}}
Traditional approaches to multitask imitation learning often rely on monolithic networks~\cite{kim_openvla_2024, octomodelteam2024octo} or language-conditioned policies~\cite{ha_scaling_2023, reussGoalConditionedImitationLearning2023a}, which limit scalability, reusability, and interpretability. While early research established modular architectures to improve task decomposition~\cite{devin2016learningmodularneuralnetwork, andreas2017modularmultitaskreinforcementlearning}, modern Sparse Diffusion Policy (SDP)~\cite{wang2024sparse_dp_moe} and variational distillation methods for MoE~\cite{zhou2024variationaldistillationdiffusionpolicies} extend this modular principle by introducing MoE layers in diffusion models, activating sparse expert sets based on observations. While this modular design enables expert reuse and policy expansion, it suffers from instability and load imbalance~\cite{lin2024moma}. Mixture-of-Denoising-Experts (MoDE)~\cite{reuss2024mode} conditions expert routing on noise level, distributing learning across noise levels, making its experts less interpretable or transferable across tasks. In contrast, \ourshort composes diffusion models through continuous score aggregation, avoiding hard expert selection and ensuring all components are jointly optimized. This promotes stable optimization, clear specialization, and better load balancing. While maintaining modular extensibility like MoE designs, \ourshort allows efficient adaptation by adding new components without overwriting prior skills.


\section{\ourshort: \textcolor{Maroon}{Factorized Diffusion Policy}}

We aim to develop a modular policy architecture that scales to diverse manipulation tasks and supports efficient adaptation to new ones. Traditional monolithic policies struggle with the complexity and multimodality of real-world action distributions, while modular alternatives like MoE suffer from training instability and poor expert interpretability. Our proposed \ourshort, which directly factorizes the policy into a set of composable diffusion models. Each component captures a distinct behavioral mode, and the final action is produced via a weighted aggregation of these modules conditioned on the current observation (Fig.~\ref{fig:our_pipeline}).

\subsection{Probabilistic Policy Modeling}

We factorize the action distribution as the product of a set of composed distributions 
\begin{equation*}
    p(\mathbf{a}_t | \mathbf{o}_t) \propto \prod_i  p_i (\mathbf{a}_{t} | \mathbf{o}_t)^{w_{t,i}},    
\end{equation*}
where $\{ w_{t,i} \}$ are observation-dependent weights associated with each component distribution. Intuitively, $p(\mathbf{a}_t | \mathbf{o}_t)$ represents the intersection (logical AND) of individual distributions, assigning high likelihood to samples commonly favored by all component distributions. Moreover, each diffusion component $p_i (\mathbf{a}_{t} | \mathbf{o}_t)$ can be interpreted as imposing a behavioral constraint (e.g., collision avoidance, precise grasping)~\cite{yang2023compositionaldiffusionbasedcontinuousconstraint}. The composed distribution thus captures the intersection of constraints, naturally framing action generation as constraint satisfaction while maintaining a probabilistic interpretation.

Denoising Diffusion Probabilistic Model (DDPM) framework~\cite{ho2020ddpm} is adopted to model each component distribution $p_i (\mathbf{a}_{t} | \mathbf{o}_t)$. To sample from each component, we start from a noisy action sample $\mathbf{a}^K_{t,i} \sim \mathcal{N}(\mathbf{0}, \mathbf{I})$, and iteratively refine it using a noise prediction network $\boldsymbol{\varepsilon}_{\theta_i}(\mathbf{a}^k_{t,i}, \mathbf{o}_t, k)$, progressively denoising over $k$ steps:
\begin{equation*}
    \mathbf{a}^{k-1}_{t,i} = \alpha_k \left( \mathbf{a}^k_{t,i} - \gamma_k \, \boldsymbol{\varepsilon}_{\theta_i}(\mathbf{a}^k_{t,i}, \mathbf{o}_t, k) + \mathcal{N}(\mathbf{0}, \sigma_k^2 \mathbf{I}) \right),
\end{equation*}
where $\alpha_k$, $\gamma_k$, and $\sigma_k$ define the noise schedule. This process closely resembles Stochastic Langevin Dynamics~\cite{welling2011bayesian}, with $\boldsymbol{\varepsilon}_{\theta_i}$ estimating the score function $\nabla \log p_i (\mathbf{a}_{t,i} | \mathbf{o}_t)$~\cite{vincent2011score_matching}. 

Training of DDPM minimizes the mean squared error (MSE) between the true added noise $\boldsymbol{\epsilon}^k$ and the network prediction:
\begin{equation}
    \mathcal{L}_\text{MSE} = \| \boldsymbol{\epsilon}^k - \boldsymbol{\varepsilon}_\theta(\mathbf{a}^0_{t,i} + \boldsymbol{\epsilon}^k, \mathbf{o}_t, k) \|^2_2, \label{equ:single_sm}
\end{equation}

where $\mathbf{a}^0_{t,i}$ is a clean trajectory sample valid under distribution $p_i (\mathbf{a}_{t} | \mathbf{o}_t)$. Minimizing this loss teaches the network to progressively denoise noisy actions conditioned on observations.

\subsection{Compositional Sampling and Routing}

We next discuss how can we sample from the actual action distribution $p (\mathbf{a}_{t} | \mathbf{o}_t)$ given DDPM formulation of component distributions $\{ p_i (\mathbf{a}_{t} | \mathbf{o}_t) \}$, as well as how to automatically discover each component distribution and optimize corresponding diffusion models jointly.

One way of viewing the composition of distributions is through the lens of energy-based models (EBM)~\cite{du2019implicitEBM}. Assume weights $\{ w_{t,i} \}$ are given, and each weighted component distribution is parameterized as $p_i (\mathbf{a}_{t} | \mathbf{o}_t) \propto e^{-w_{t,i} E_i}$, then the actual action distribution can be expressed as $p (\mathbf{a}_{t} | \mathbf{o}_t) \propto e^{-\sum_i w_{t,i} E_i}$~\cite{du2019implicitEBM}. Therefore, iterative sampling can be performed via Langevin dynamics:
\begin{equation}
    \mathbf{a}^{k-1}_t = \mathbf{a}^k_t- \gamma_k \sum_i w_{t,i} \, \nabla_{\mathbf{a}^k_t} E_i(\mathbf{a}^k_t, \mathbf{o}_t) + \xi_k, \label{equ:comp_sampling}
\end{equation}
where $\gamma_k$ controls the step size and $\xi_k$ introduces Gaussian noise. Note that we can bridge EBM with score-matching diffusion models~\cite{wang2024poco, du2024reducereuserecyclecompositional, liu2023compositionalvisualgenerationcomposable, su2024compositionalimagedecompositiondiffusion}, which updates Equ.~\ref{equ:comp_sampling} as
\begin{equation*}
    \mathbf{a}^{k-1}_t = \mathbf{a}^k_t- \gamma_k \sum_i w_{t,i} \, \boldsymbol{\varepsilon}_{\theta_i}(\mathbf{a}^k_t, \mathbf{o}_t, k) + \xi_k.
\end{equation*}

To optimize diffusion components jointly, we update MSE loss in Equ.~\ref{equ:single_sm} to
\begin{equation*}
    \mathcal{L}_\text{MSE} = \| \boldsymbol{\epsilon}^k - \sum_i w_{t,i} \, \boldsymbol{\varepsilon}_{\theta_i}(\mathbf{a}^0_t + \boldsymbol{\epsilon}^k, \mathbf{o}_t, k) \|^2_2,
\end{equation*}
where $\mathbf{a}^0_t$ is a demonstration trajectory sample. Then all diffusion components are optimized jointly end-to-end.

The weights $\{ w_{t,i} \}$ are predicted by a lightweight observation-conditioned multi-layer perceptron (MLP), referred to as \textit{router}, which is optimized along with other diffusion components. This brings the last piece of \ourshort architecture. The pseudocode for training and inference are provided in Algo.~\ref{algo:fdp_train} and Algo.~\ref{algo:fdp_inference}.

Compared to discrete MoE routing, our compositional approach avoids routing instability and expert imbalance~\cite{lin2024moma} by assigning continuous, observation-dependent weights to all components, rather than selecting a hard subset. In MoE, only a few experts are activated at each step, which can lead to underutilization of some experts and overfitting or saturation in others, especially when routing distributions are sharp or poorly calibrated. In contrast, our method aggregates contributions from all components via soft score-weighted composition, ensuring all modules remain active during optimization. Additionally, because all components participate in every training step, they receive gradient signals consistently, which encourages functional specialization.

\begin{figure}[t]
\centering
\begin{minipage}[t]{0.49\textwidth}
\begin{algorithm}[H]
\footnotesize
\caption{\ourshort Training}
\label{algo:fdp_train}
\begin{algorithmic}[1]
\setlength{\baselineskip}{\baselineskip}
\REQUIRE Dataset $\mathcal{D}$, Denoisers $\{ \boldsymbol{\varepsilon}_{\theta_i} \}$, \textsc{Router}$_\psi$
\WHILE{not converged}
    \STATE Sample $(\mathbf{a}, \mathbf{o}) \sim \mathcal{D}$ and noise $\boldsymbol{\epsilon}^k$
    \STATE $\{w_i\} \gets$ \textsc{Router}$_\psi (\mathbf{o})$
    \STATE $\mathcal{L} \gets \left\| \boldsymbol{\epsilon}^k - \sum_i w_i \, \boldsymbol{\varepsilon}_{\theta_i}(\mathbf{a} + \boldsymbol{\epsilon}^k, \mathbf{o}, k) \right\|^2_2$
    \STATE $\forall i, \theta_i \gets \theta_i + \nabla_{\theta_i} \mathcal{L}$
    \STATE $\psi \gets \psi + \nabla_\psi \mathcal{L}$
\ENDWHILE
\RETURN $\{ \boldsymbol{\varepsilon}_{\theta_i} \}$
\end{algorithmic}
\end{algorithm}
\end{minipage}
\hfill
\begin{minipage}[t]{0.49\textwidth}
\begin{algorithm}[H]
\footnotesize
\caption{\ourshort Inference}
\label{algo:fdp_inference}
\begin{algorithmic}[1]
\setlength{\baselineskip}{\baselineskip}
\REQUIRE Denoisers $\{ \boldsymbol{\varepsilon}_i \}$, \textsc{Router}, Observation $\mathbf{o}_t$
\STATE $\{w_{t,i}\} \gets$ \textsc{Router}$ (\mathbf{o}_t)$
\STATE $\mathbf{a}_t^K \gets \mathcal{N}(\mathbf{0}, \mathbf{I})$
\FOR{$k \gets K, K-1, ..., 1$}
    \STATE $\nabla \mathbf{a}^k \gets \sum_i w_{t,i} \, \boldsymbol{\varepsilon}_i (\mathbf{a}_t^k, \mathbf{o}_t, k)$
    \STATE $\mathbf{a}_t^{k-1} \gets \mathbf{a}_t^k - \gamma_k \nabla \mathbf{a}^k + \mathcal{N}(\mathbf{0}, \sigma_k \mathbf{I})$
\ENDFOR
\STATE $\mathbf{a}_t \gets \mathbf{a}_t^0$
\RETURN $\mathbf{a}_t$
\end{algorithmic}
\end{algorithm}
\end{minipage}
\end{figure}

\subsection{Multitask Learning and Adaptation}

\subsubsection{\textbf{Multitask Learning}}
This factorization is particularly well-suited for multitask imitation learning, where action distributions are inherently multimodal due to diverse object properties, contact dynamics, and task goals. In contrast to monolithic policies that must capture all modes simultaneously, \ourshort distributes complexity evenly across diffusion components, each modeling a coherent subspace of behaviors. Unlike MoE policies, where skills may span combinations of experts across layers, our formulation yields disentangled sub-skills.

\subsubsection{\textbf{Adapting to New Tasks}}
The modularity of \ourshort also enables efficient adaptation to unseen tasks. Instead of retraining the full model, we adapt by introducing a new diffusion component $\boldsymbol{\varepsilon}_{\theta_{\text{new}}}$, initialized via \textit{upcycling}~\cite{lin2024moma} -- copying weights from existing components. The updated score function becomes:
\begin{equation*}
    \boldsymbol{\varepsilon}_{\text{adapt}}(\mathbf{a}_t^k, \mathbf{o}_t, k) = \sum_i w_i \, \boldsymbol{\varepsilon}_{\theta_i}(\mathbf{a}_t^k, \mathbf{o}_t, k) + w_{\text{new}} \, \boldsymbol{\varepsilon}_{\theta_{\text{new}}}(\mathbf{a}_t^k, \mathbf{o}_t, k),
\end{equation*}
where only $\boldsymbol{\varepsilon}_{\theta_{\text{new}}}$ and the new router are updated during adaptation, using the training loss in Equ.~\ref{equ:single_sm}. All previously trained components $\{ \boldsymbol{\varepsilon}_{\theta_i} \}$ are frozen. Freezing existing components ensures that the optimization focuses solely on capturing novel task dynamics without disrupting existing capabilities, thereby mitigating catastrophic forgetting. Such selective adaptation significantly reduces the number of trainable parameters and the amount of supervision required. In contrast, MoE models, where overlapping expert roles make modular reuse and analysis more difficult.

Finally, \ourshort supports heterogeneous architectures -- diffusion models can vary in architecture and size, enabling scalable allocation of computation to match task complexity. This extensibility makes \ourshort broadly applicable in diverse and evolving robotic domains.


\section{Experiments}

In this section, we aim to empirically investigate several key questions regarding our proposed policy architecture: (1) Whether factorizing the complex action distribution into simpler distributions captured by smaller diffusion models can improve overall policy learning and performance. (2) Whether the modular structure of \ourshort, composed of multiple diffusion-based expert modules, facilitates more efficient and effective task transfer and adaptation. (3) How different adaptation strategies compare, highlighting trade-offs such as data efficiency, policy performance, and compute.

\subsection{Experiments Setup}

\begin{figure}[t]
    \centering
    \begin{subfigure}{.38\columnwidth}
        \includegraphics[width=\columnwidth]{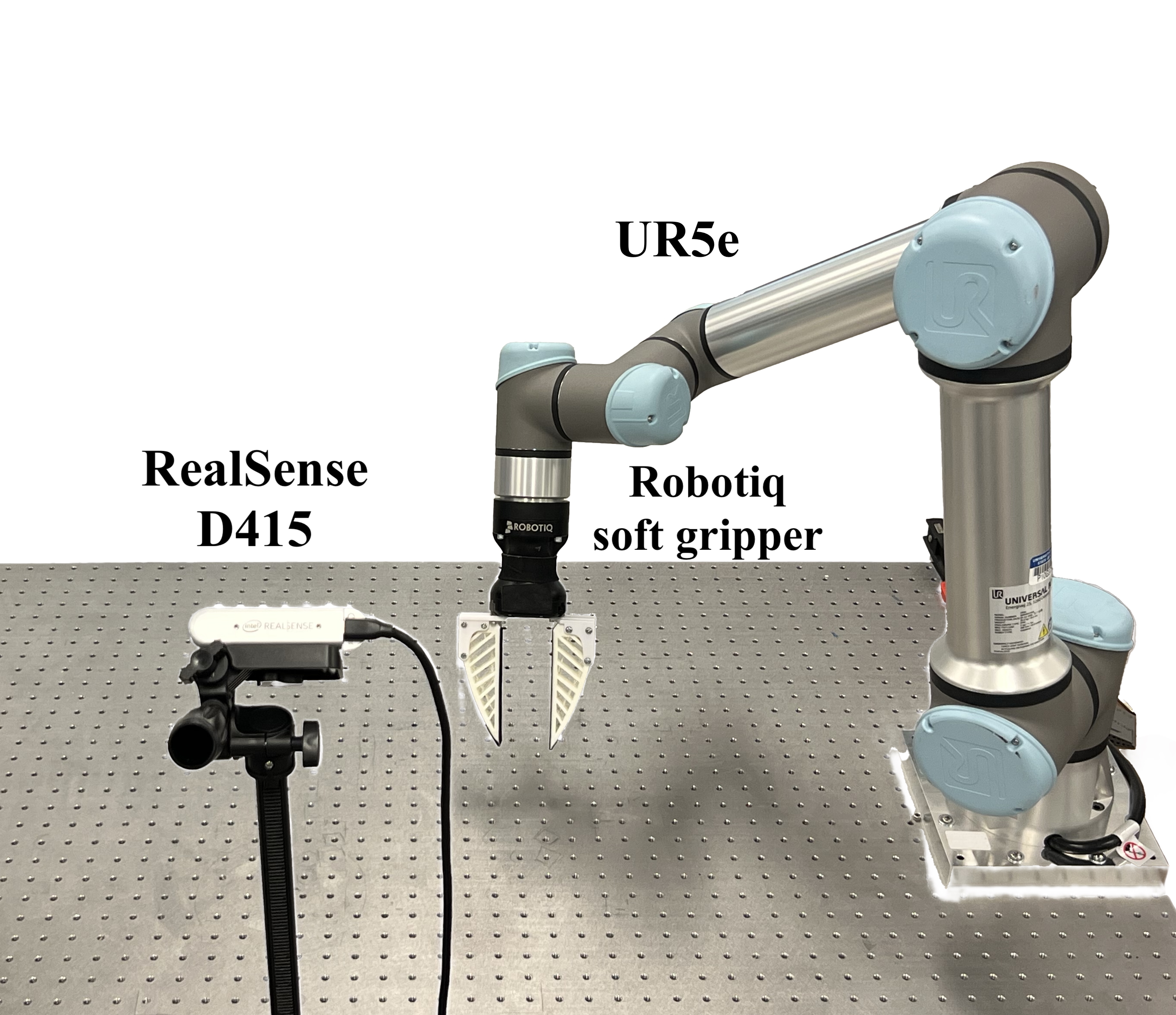}
        \caption{Real-world workspace.}
        \label{fig:real_workspace_setup}
    \end{subfigure}
    \hfill
    \begin{subfigure}{.28\columnwidth}
        \includegraphics[width=\columnwidth]{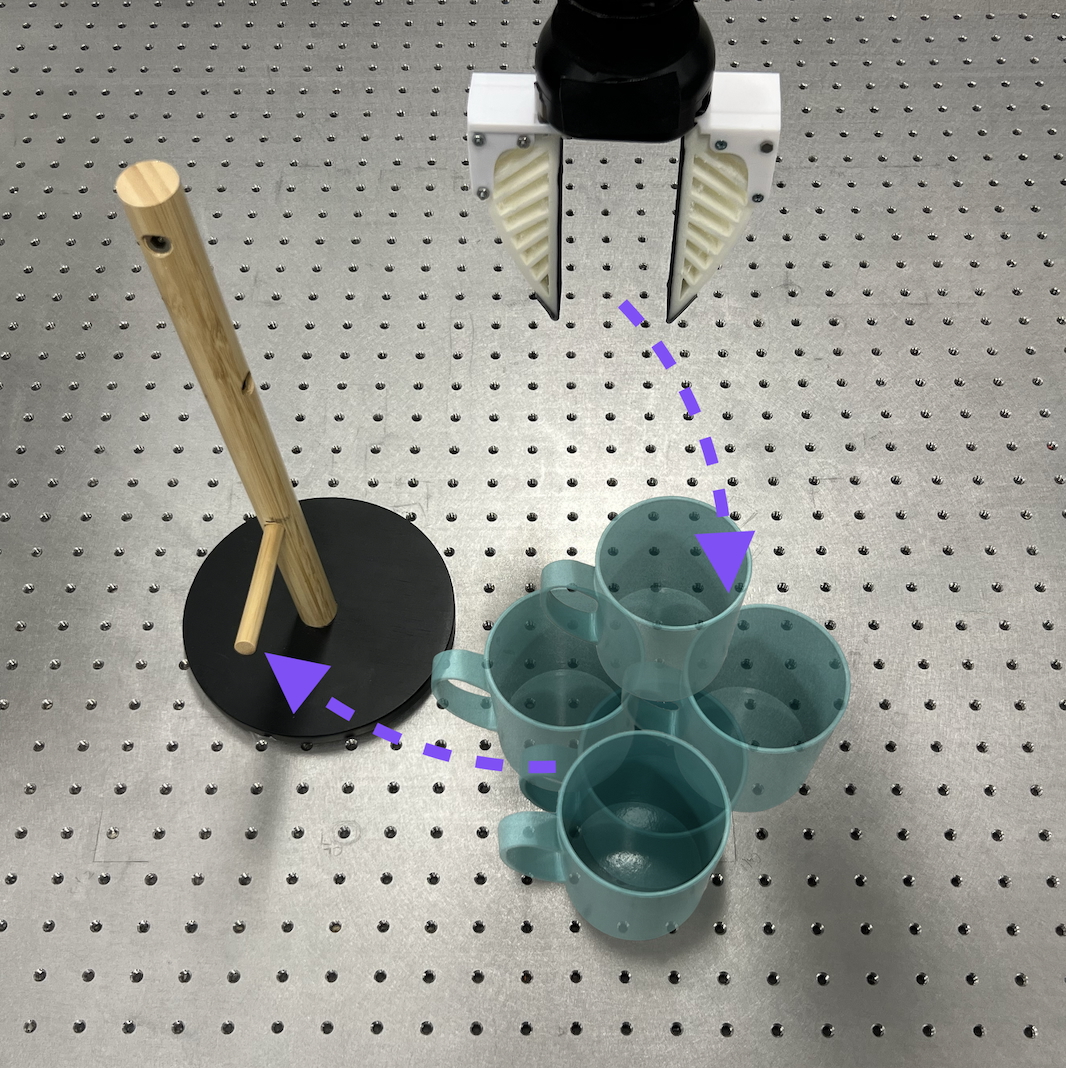}
        \caption{\textit{hang-X}}
    \end{subfigure}
    \hfill
    \begin{subfigure}{.28\columnwidth}
        \includegraphics[width=\columnwidth]{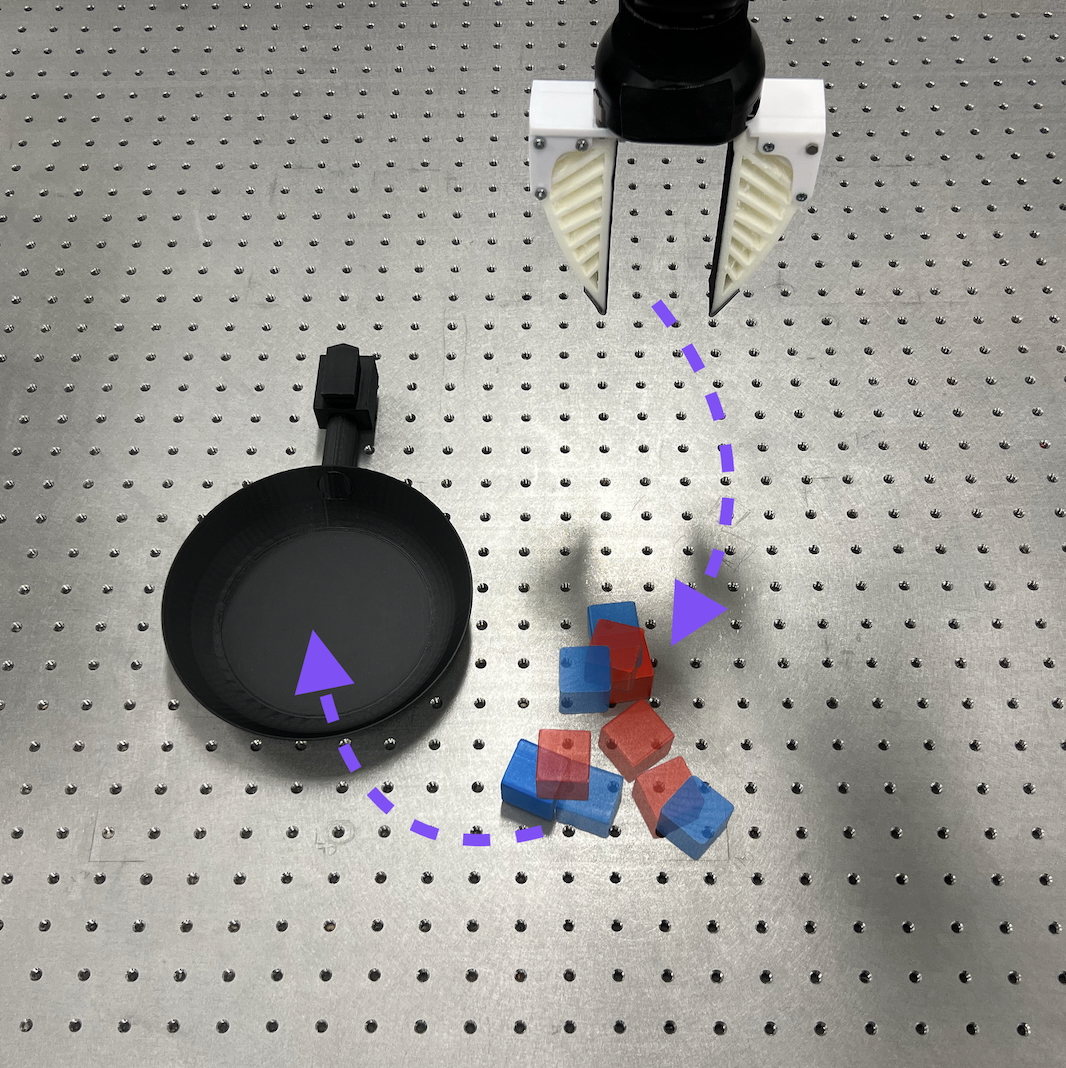}
        \caption{\textit{cube-X}}
    \end{subfigure}
    \caption{\textbf{Real-world setup and task illustrations}. (a) Workspace setup with a UR5e arm, Robotiq gripper, and RealSense D415 camera. (b) High-level task illustrations.}
    \label{fig:real_overview}
\end{figure}

\begin{figure*}[t]
\centering
\begin{minipage}[t]{0.29\textwidth} 
    \centering
    \begin{subfigure}[t]{.94\linewidth}
        \includegraphics[width=\linewidth]{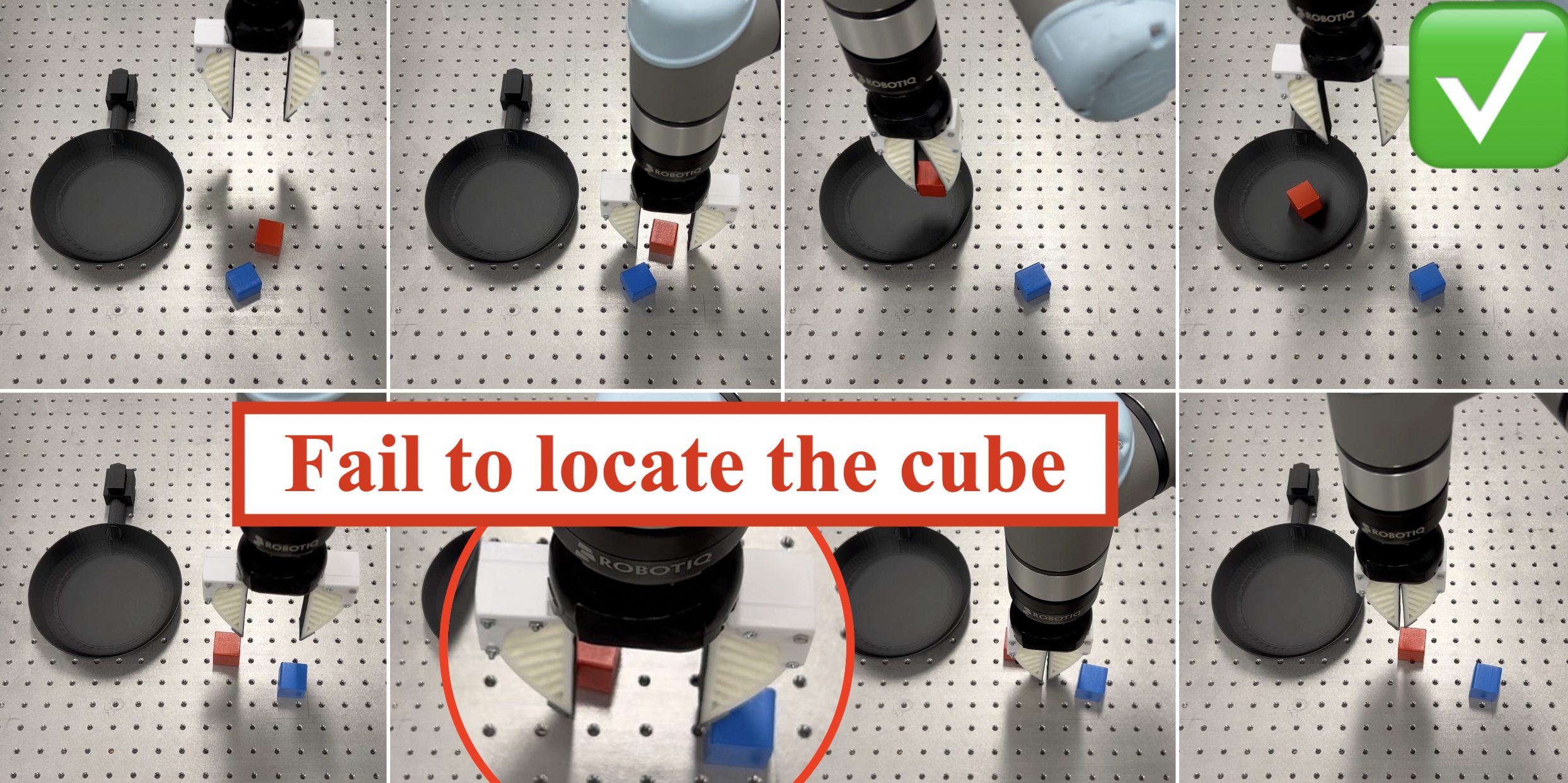}
    \end{subfigure}
    \hfill
    \begin{subfigure}[t]{.94\linewidth}
        \includegraphics[width=\linewidth]{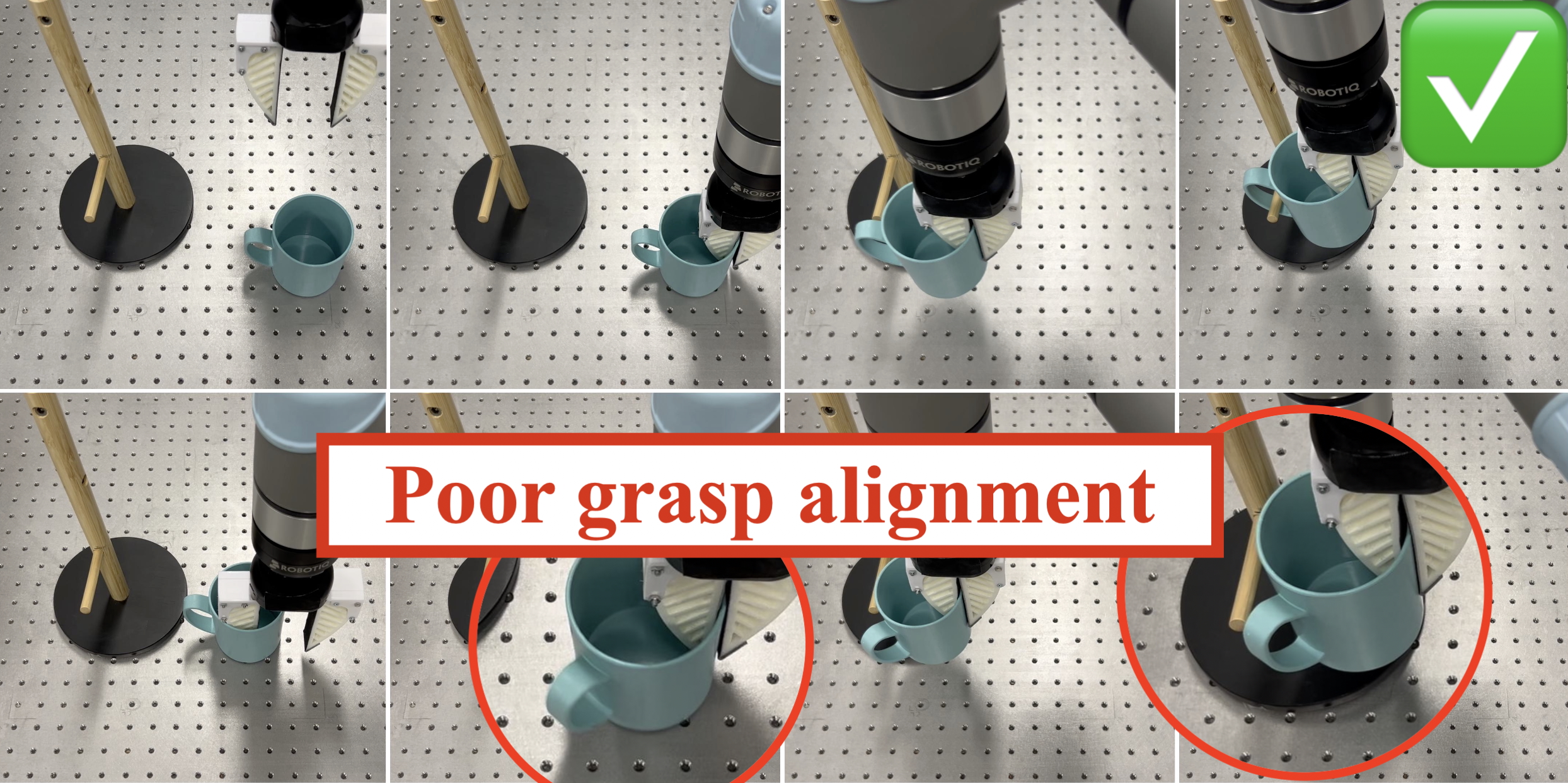}
    \end{subfigure}
    \caption{\textbf{Real-world rollouts.} Top: \textit{cube-X}. Bottom: \textit{hang-X}. Top and bottom rows show success cases and baseline failure modes.}
    \label{fig:real_qualitative_comparison}
    \vspace{-10pt}
\end{minipage}
\hfill
\begin{minipage}[t]{0.7\textwidth}  
\centering
\scriptsize
\setlength{\tabcolsep}{3.5pt}
\begin{tabular}{l|ccccccccc}
\toprule
\multicolumn{10}{c}{\textbf{MetaWorld}} \\
 & & Door & Drawer & & Window & Peg \\
Policy & & Open & Open & Assembly & Close & Insert & Hammer & & Avg. \\
\midrule
DP     &   & \phantom{0}87.0 {\tiny ± 1.92} & 100.0 {\tiny ± 0.00} & 100.0 {\tiny ± 0.00} & \phantom{0}94.0 {\tiny ± 0.89} & 20.0 {\tiny ± 1.22} & 24.0 {\tiny ± 0.89} & & 70.8 {\tiny ± 0.24} \\
SDP     &  & \phantom{0}80.0 {\tiny ± 1.87} & 100.0 {\tiny ± 0.00} & 100.0 {\tiny ± 0.00} & 100.0 {\tiny ± 0.00} & 20.5 {\tiny ± 0.84} & 18.0 {\tiny ± 1.64} & & 69.8 {\tiny ± 0.51} \\
MoDE     & & 100.0 {\tiny ± 0.00} & 100.0 {\tiny ± 0.00} & \phantom{0}94.0 {\tiny ± 2.51} & 100.0 {\tiny ± 0.00} & 19.5 {\tiny ± 2.77} & 23.5 {\tiny ± 0.89} & & 72.8 {\tiny ± 0.76} \\
\ourshort & & 100.0 {\tiny ± 0.00} & 100.0 {\tiny ± 0.00} & 100.0 {\tiny ± 0.00} & 100.0 {\tiny ± 0.00} & 26.5 {\tiny ± 2.19} & 22.0 {\tiny ± 2.39} & & 74.8 {\tiny ± 0.67} \\
\bottomrule
\addlinespace[5pt]
\toprule
\multicolumn{10}{c}{\textbf{RLBench}} \\
 & Close & Close & Close & Close & Toilet Seat & Take & Reach & Pick \& &  \\
Policy & Box & Drawer & Fridge & Microwave & Down & Umbrella & Target & Lift & Avg. \\
\midrule
DP        & 33.5 {\tiny ± 1.52} & 100.0 {\tiny ± 0.00} & 77.0 {\tiny ± 1.64} & 65.5 {\tiny ± 1.48} & 64.0 {\tiny ± 2.19} & 19.0 {\tiny ± 1.95} & \phantom{0}3.5 {\tiny ± 0.89} & 2.0 {\tiny ± 0.84} & 45.6 {\tiny ± 0.43} \\
SDP       & 24.0 {\tiny ± 1.34} & 100.0 {\tiny ± 0.00} & 68.0 {\tiny ± 3.11} & 42.5 {\tiny ± 2.00} & 58.5 {\tiny ± 2.88} & \phantom{0}6.5 {\tiny ± 1.14} & 36.0 {\tiny ± 2.61} & 1.0 {\tiny ± 0.55} & 42.1 {\tiny ± 0.70} \\
MoDE      & 31.0 {\tiny ± 2.88} & \phantom{0}97.0 {\tiny ± 2.17} & 70.5 {\tiny ± 1.64} & 50.5 {\tiny ± 1.10} & 60.0 {\tiny ± 1.73} & 10.5 {\tiny ± 1.92} & 37.0 {\tiny ± 1.10} & 3.5 {\tiny ± 1.95} & 45.0 {\tiny ± 0.87} \\
\ourshort & 72.0 {\tiny ± 1.30} & 100.0 {\tiny ± 0.00} & 75.5 {\tiny ± 1.10} & 93.0 {\tiny ± 1.92} & 27.5 {\tiny ± 2.12} & 78.0 {\tiny ± 2.17} & 61.0 {\tiny ± 1.52} & 4.5 {\tiny ± 2.39} & 63.9 {\tiny ± 0.23} \\
\bottomrule
\end{tabular}
\vspace{0.5pt}
\captionof{table}{\textbf{Multitask learning evaluation on MetaWorld and RLBench}. Results report the mean and standard error over 5 seeds, with 40 samples evaluated per seed.}
\label{tab:sim_pretrain}
\end{minipage}

\end{figure*}

\begin{table*}[t]
\centering
\scriptsize
\setlength{\tabcolsep}{6.0pt}
\begin{tabular}{cl|ccccc|ccccc}
\toprule
                     &                                & \multicolumn{5}{c}{\textbf{MetaWorld}}                     & \multicolumn{5}{c}{\textbf{RLBench}} \\
 & \multicolumn{1}{c|}{} & Door & Drawer & & Window & \multicolumn{1}{c|}{} & Open & Open & Open & Open & \\
Method & \multicolumn{1}{c|}{Policy} & Close & Close & Disassemble & Open & \multicolumn{1}{c|}{Avg.} & Box & Drawer & Fridge & Microwave & Avg. \\ \midrule
\multirow{4}{*}{\begin{tabular}{@{}c@{}}
Full \\ Parameter\end{tabular}}  
    & \multicolumn{1}{c|}{DP}        
    & 100.0 {\tiny ± 0.00} & 100.0 {\tiny ± 0.00} & 60.5 {\tiny ± 1.10} & 100.0 {\tiny ± 0.00} & \multicolumn{1}{c|}{90.1 {\tiny ± 0.27}} 
    & 75.5 {\tiny ± 1.30} & 81.0 {\tiny ± 0.55} & 64.0 {\tiny ± 2.07} & 63.5 {\tiny ± 1.52} & 71.0 {\tiny ± 0.76} \\

    & \multicolumn{1}{c|}{SDP}       
    & \phantom{0}89.5 {\tiny ± 2.05} & 100.0 {\tiny ± 0.00} & 58.0 {\tiny ± 2.17} & 100.0 {\tiny ± 0.00} & \multicolumn{1}{c|}{86.9 {\tiny ± 0.31}} 
    & 67.0 {\tiny ± 0.84} & 88.5 {\tiny ± 2.88} & 58.5 {\tiny ± 2.61} & 66.0 {\tiny ± 1.52} & 70.0 {\tiny ± 0.90} \\

    & \multicolumn{1}{c|}{MoDE}      
    & 100.0 {\tiny ± 0.00} & 100.0 {\tiny ± 0.00} & 66.5 {\tiny ± 2.97} & 100.0 {\tiny ± 0.00} & \multicolumn{1}{c|}{91.6 {\tiny ± 0.74}} 
    & 60.5 {\tiny ± 0.45} & 88.0 {\tiny ± 1.92} & 71.0 {\tiny ± 0.89} & 75.0 {\tiny ± 1.22} & 73.6 {\tiny ± 0.48} \\

    & \multicolumn{1}{c|}{\ourshort} 
    & 100.0 {\tiny ± 0.00} & 100.0 {\tiny ± 0.00} & 62.0 {\tiny ± 2.17} & 100.0 {\tiny ± 0.00} & \multicolumn{1}{c|}{90.5 {\tiny ± 0.54}} 
    & 87.0 {\tiny ± 1.92} & 72.5 {\tiny ± 3.39} & 65.0 {\tiny ± 1.87} & 79.0 {\tiny ± 1.67} & 75.9 {\tiny ± 0.91} \\ \midrule

\multirow{3}{*}{Router}       
    & \multicolumn{1}{c|}{SDP}       
    & \phantom{0}72.0 {\tiny ± 1.30} & \phantom{00}0.0 {\tiny ± 0.00} & \phantom{0}1.0 {\tiny ± 0.55} & \phantom{00}4.0 {\tiny ± 1.52} & \multicolumn{1}{c|}{19.3 {\tiny ± 0.67}} 
    & \phantom{0}4.5 {\tiny ± 1.92} & 23.5 {\tiny ± 1.95} & 11.5 {\tiny ± 1.14} & 15.5 {\tiny ± 2.49} & 13.8 {\tiny ± 0.40} \\

    & \multicolumn{1}{c|}{MoDE}      
    & \phantom{00}4.0 {\tiny ± 2.30} & \phantom{00}1.0 {\tiny ± 0.89} & \phantom{0}1.5 {\tiny ± 1.34} & \phantom{0}24.0 {\tiny ± 1.67} & \multicolumn{1}{c|}{\phantom{0}7.6 {\tiny ± 1.08}} 
    & \phantom{0}2.5 {\tiny ± 1.41} & \phantom{0}4.0 {\tiny ± 1.95} & 19.5 {\tiny ± 0.84} & \phantom{0}1.5 {\tiny ± 1.34} & \phantom{0}6.9 {\tiny ± 0.71} \\

    & \multicolumn{1}{c|}{\ourshort} 
    & \phantom{0}85.5 {\tiny ± 1.48} & \phantom{0}64.5 {\tiny ± 2.28} & \phantom{0}4.0 {\tiny ± 2.19} & \phantom{00}5.5 {\tiny ± 0.84} & \multicolumn{1}{c|}{39.9 {\tiny ± 0.67}} 
    & 17.5 {\tiny ± 1.41} & \phantom{0}1.5 {\tiny ± 1.34} & 17.5 {\tiny ± 1.73} & \phantom{0}0.0 {\tiny ± 0.00} & \phantom{0}9.1 {\tiny ± 0.82} \\ \midrule

\multirow{3}{*}{\begin{tabular}{@{}c@{}}+ Observation \\ Encoder\end{tabular}}  
    & \multicolumn{1}{c|}{SDP}       
    & 100.0 {\tiny ± 0.00} & 100.0 {\tiny ± 0.00} & 31.0 {\tiny ± 0.89} & 100.0 {\tiny ± 0.00} & \multicolumn{1}{c|}{82.8 {\tiny ± 0.22}} 
    & 25.5 {\tiny ± 0.84} & 48.0 {\tiny ± 2.17} & 41.0 {\tiny ± 1.14} & 22.5 {\tiny ± 1.87} & 34.3 {\tiny ± 0.45} \\

    & \multicolumn{1}{c|}{MoDE}      
    & \phantom{0}93.0 {\tiny ± 0.45} & 100.0 {\tiny ± 0.00} & 40.0 {\tiny ± 1.58} & 100.0 {\tiny ± 0.00} & \multicolumn{1}{c|}{83.2 {\tiny ± 0.37}} 
    & 19.0 {\tiny ± 1.34} & 65.5 {\tiny ± 1.10} & 29.5 {\tiny ± 1.30} & 25.0 {\tiny ± 1.00} & 34.8 {\tiny ± 0.91} \\

    & \multicolumn{1}{c|}{\ourshort} 
    & 100.0 {\tiny ± 0.00} & \phantom{0}93.0 {\tiny ± 2.68} & 52.5 {\tiny ± 1.87} & 100.0 {\tiny ± 0.00} & \multicolumn{1}{c|}{86.4 {\tiny ± 0.99}} 
    & 30.0 {\tiny ± 3.08} & 57.0 {\tiny ± 1.48} & 47.5 {\tiny ± 2.55} & 18.5 {\tiny ± 1.82} & 38.3 {\tiny ± 1.19} \\ \midrule

\multirow{3}{*}{\begin{tabular}{@{}c@{}}+ New \\ Module\end{tabular}} 
    & \multicolumn{1}{c|}{SDP}       
    & 100.0 {\tiny ± 0.00} & 100.0 {\tiny ± 0.00} & 60.5 {\tiny ± 1.79} & 100.0 {\tiny ± 0.00} & \multicolumn{1}{c|}{90.1 {\tiny ± 0.45}} 
    & 69.0 {\tiny ± 1.14} & 23.5 {\tiny ± 2.88} & 40.5 {\tiny ± 1.92} & 22.5 {\tiny ± 1.00} & 38.9 {\tiny ± 0.89} \\

    & \multicolumn{1}{c|}{MoDE}      
    & 100.0 {\tiny ± 0.00} & 100.0 {\tiny ± 0.00} & 61.5 {\tiny ± 2.30} & 100.0 {\tiny ± 0.00} & \multicolumn{1}{c|}{90.4 {\tiny ± 0.58}} 
    & 58.0 {\tiny ± 1.92} & 54.0 {\tiny ± 1.14} & 58.0 {\tiny ± 1.79} & 53.0 {\tiny ± 1.92} & 55.8 {\tiny ± 0.78} \\

    & \multicolumn{1}{c|}{\ourshort} 
    & 100.0 {\tiny ± 0.00} & 100.0 {\tiny ± 0.00} & 68.5 {\tiny ± 1.52} & 100.0 {\tiny ± 0.00} & \multicolumn{1}{c|}{92.1 {\tiny ± 0.38}} 
    & 86.5 {\tiny ± 1.34} & 77.0 {\tiny ± 1.10} & 78.0 {\tiny ± 2.28} & 77.5 {\tiny ± 3.08} & 79.8 {\tiny ± 0.91} \\ 
\bottomrule
\end{tabular}
\vspace{+1.5pt}
\caption{\textbf{Adaptation evaluation on MetaWorld and RLBench}. We pretrain policies on tasks shown in Table~\ref{tab:sim_pretrain}, then adapt them to these successive tasks. We report mean and standard error over 5 seeds, with 60 MetaWorld samples and 40 RLBench samples per seed.}
\label{tab:sim_adaptation}
\vspace{-20pt}
\end{table*}

We evaluate policies on 30+ tasks in simulation across MetaWorld~\cite{yu2021metaworldbenchmark}, RLBench~\cite{james2019rlbenchbenchmark}, and LIBERO~\cite{liu2023libero}. Real-world experiments use a UR5e arm with a Robotiq gripper and a RealSense D415 camera (Fig.~\ref{fig:real_workspace_setup}). We evaluate policies on 4 distinct tasks: \textit{cube red}, \textit{cube blue}, \textit{hang low}, and \textit{hang high}. The tasks \textit{cube-X} involve picking up a cube of color \textit{X} from the tabletop and placing it into a designated bowl. The \textit{hang-X} tasks require the robot to grasp a mug from the tabletop and precisely hang it on the \textit{X} branch of a mug stand positioned on the table. Illustrations and setups of these real-world tasks are shown in Fig.~\ref{fig:real_overview}. For MetaWorld and RLBench, goals are implicitly specified by the scene configuration, whereas for LIBERO and real-world experiments, tasks are identified using discrete integer task indices.

\subsection{Implementations}
\label{sec:implementation}

All policies take RGB images and joint angles as input and predict absolute joint angle trajectories. A history window of size 2 is used, with 16-step trajectories predicted and 8 steps executed. While we employ DDPM~\cite{ho2020ddpm}, \ourshort is solver-agnostic; alternative samplers like DDIM~\cite{song2022ddim} offer comparable performance with reduced inference latency. We compare \ourshort against three baselines: DP~\cite{chi2024diffusionpolicy}, a monolithic diffusion policy; SDP~\cite{wang2024sparse_dp_moe}, a MoE-based diffusion policy with observation-conditioned routing; and MoDE~\cite{reuss2024mode}, a MoE variant with routing based on noise levels. We follow the original configurations used in each baselines, and proportionally reduce the model size of MoDE to match others. We refer readers to the original papers for more details on architecture and training. In \ourshort, four U-Net diffusion modules are composed. For adaptation, we adopt the upcycling strategy~\cite{lin2024moma} to initialize new MoE experts or diffusion components from existing ones.

\subsection{Multitask Learning}

We first investigate whether decomposing complex motion distributions into simpler, behavior-specialized components can improve policy performance in multitask settings.

\subsubsection{\textbf{Simulation}} 
We evaluate \ourshort on 6 MetaWorld tasks (25 demonstrations each) and 8 RLBench tasks (50 demonstrations each). All methods are evaluated over 40 rollouts per task, with results shown in Table~\ref{tab:sim_pretrain}. The DP baseline performs surprisingly well, particularly on tasks like \textit{drawer open}, \textit{assembly}, and \textit{hammer}, which primarily involve reaching and grasping and exhibit fewer multimodal behaviors -- making them easier to solve with a single model. Among modular baselines, SDP underperforms due to instability common in training MoE architectures~\cite{lin2024moma}: too few experts limit expressiveness, while too many can cause overfitting and noisy routing. MoDE performs reasonably by routing based on the noise level, but still inherits instability from MoE training~\cite{reuss2024mode}. In contrast, \ourshort's compositional structure avoids abrupt routing decisions by continuously composing diffusion component outputs via score-weighted aggregation, which enables stable training and more balanced component specialization of the multimodal action distributions.

\begin{wraptable}{r}{0.465\columnwidth}
    \vspace{-10pt}
    \centering
    \scriptsize
    \setlength{\tabcolsep}{5.5pt}
    \begin{tabular}{l|cc}
    \toprule
    Policy    & Cube Red & Hang Low \\ \midrule
    DP        & 14 / 20    & 16 / 20  \\
    SDP       & 15 / 20    & 13 / 20  \\
    MoDE      & 14 / 20    & 16 / 20 \\
    \ourshort & \textbf{15 / 20}    & \textbf{17 / 20} \\
    \bottomrule
    \end{tabular}
    \captionof{table}{\textbf{Real-world multitask success rates.} Average over 20 trials. Tasks: \textit{cube red} and \textit{hang low}.}
    \label{tab:real_pretrain}
    \vspace{-15pt}
\end{wraptable}

\subsubsection{\textbf{Real-world}}
We further evaluate our method in real-world settings on two tasks: \textit{cube red} (300 demonstrations) and \textit{hang low} (200 demonstrations). 20 samples are used for evaluation, and results are summarized in Table~\ref{tab:real_pretrain}. The DP baseline often overfits to specific joint trajectories, failing to attend to RGB inputs due to the multimodal and perceptually complex nature of the tasks. By contrast, \ourshort captures diverse behavior patterns more effectively by decomposing the action distribution across sub-modules. This results in higher success rates. Fig.~\ref{fig:real_qualitative_comparison} shows qualitative failure cases from baseline methods, which struggle to capture the complex distribution, resulting in imprecise end-effector poses and frequent task failures.

\subsection{Task Transfer and Adaptation}

In this section, we evaluate the adaptability of \ourshort in adapting to novel tasks under limited data. We first pretrain policies on the tasks shown in Table~\ref{tab:sim_pretrain}, then evaluate
adaptation performance on new tasks shown in Table~\ref{tab:sim_adaptation}. We compare several adaptation strategies: full-parameter fine-tuning, partial fine-tuning of the router, observation encoder, and selective module expansion via new expert components. The proportion of tunable parameters of \ourshort under different adaptation strategies are (a) \textit{router-only} activates 0.5\%, (b)\textit{+ observation encoder} activates 11\%, and (c)\textit{+ new module} activates 27\% of parameters.

\begin{table}[t]
\centering
\scriptsize
\setlength{\tabcolsep}{10.0pt}
\begin{tabular}{l|lcc}
\toprule
Method & Policy & Cube Blue & Hang High\\ \midrule
\multirow{4}{*}{Full Param.} & DP & 15 / 20 & 17 / 20  \\
                             & SDP & 14 / 20 & 16 / 20 \\
                             & MoDE & 15 / 20 & 16 / 20  \\
                             & \ourshort & \textbf{17 / 20} & 15 / 20 \\ \midrule
\multirow{3}{*}{Router + Obs. Enc.} & SDP & 10 / 20 & \phantom{0}9 / 20  \\
                                    & MoDE & 10 / 20 & 11 / 20  \\
                                    & \ourshort & \textbf{11 / 20} & \textbf{11 / 20} \\ \midrule
\multirow{3}{*}{+ New Module} & SDP & 13 / 20 & 11 / 20 \\
                              & MoDE & 14 / 20 & 13 / 20 \\
                              & \ourshort & \textbf{17 / 20} & \textbf{17 / 20} \\ \bottomrule
\end{tabular}
\captionof{table}{\textbf{Adaptation in real-world}. Evaluated on \textit{cube blue} and \textit{hang high}. Pretrained on \textit{cube red} and \textit{hang low}.}
\label{tab:real_adaptation}
\end{table}

\subsubsection{\textbf{Simulation}}
We evaluate adaptation performance on 4 MetaWorld tasks and 4 RLBench tasks, using 10 and 25 demonstrations per task, respectively. We run 60 evaluations for MetaWorld and 40 evaluations for RLBench. As shown in Table~\ref{tab:sim_adaptation}, full-parameter fine-tuning achieves strong performance but is computationally intensive. Partial fine-tuning -- modifying only the router or including the observation encoder -- offers limited gains. In contrast, adding new modules (two expert blocks per layer for MoE-based methods and a new diffusion component for \ourshort) consistently improves performance. \ourshort benefits most from this strategy, leveraging its compositional structure to reuse prior knowledge while efficiently learning new behaviors. 

\subsubsection{\textbf{Real-world}}
We further evaluate adaptation on two real-world tasks, each with 100 demonstrations. 20 samples are used for evaluation. Results in Table~\ref{tab:real_adaptation} echo the simulation trends. While full-parameter fine-tuning performs reasonably well, it is resource-intensive. Partial fine-tuning yields modest improvements. The most effective strategy across all methods involves introducing new modules. Under this setting, \ourshort achieves the best performance, highlighting the advantage of its modular design for rapid and robust adaptation even in complex, real-world scenarios.

\begin{figure}[t]
\centering
\begin{minipage}[t]{0.45\linewidth} 
    \centering
    \includegraphics[width=\linewidth]{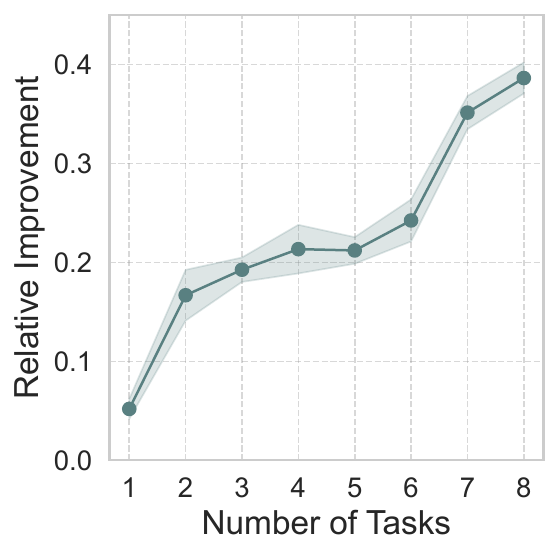}
\end{minipage}
\hfill
\begin{minipage}[t]{0.5\linewidth} 
    \vspace{-105pt}
    \captionof{figure}{\textbf{Relative success rate improvement of \ourshort over DP.} \ourshort~'s advantage increases as the number of tasks grows. Selected from RLBench. We report mean and standard error over 5 seeds.}
    \label{fig:num_task_scaling}
\end{minipage}
\end{figure}

\begin{table}[t]
\centering
\begin{minipage}[t]{0.45\linewidth}  
    \centering
    \scriptsize
    \setlength{\tabcolsep}{4.5pt}
    \begin{tabular}{c|cc}
    \toprule
    \# Comp & MetaWorld & RLBench \\ \midrule
    2  & 86.7 {\tiny ± 0.89} & 54.4 {\tiny ± 1.22}   \\
    3  & 90.0 {\tiny ± 0.71} & 58.8 {\tiny ± 0.84}  \\
    4  & 91.3 {\tiny ± 0.45} & 63.9 {\tiny ± 0.23}   \\
    5  & 91.3 {\tiny ± 0.89} & 64.4 {\tiny ± 0.71}  \\
    6  & 91.7 {\tiny ± 1.14} & 65.0 {\tiny ± 0.45}  \\
    7  & 91.9 {\tiny ± 0.55} & 65.6 {\tiny ± 0.84}  \\ 
    \bottomrule
    \end{tabular}
\end{minipage}
\hfill
\begin{minipage}[t]{0.5\linewidth}  
    \vspace{-35pt}
    \captionof{table}{\textbf{Multitask performance of \ourshort with different numbers of components.} Performance improves up to 4 components and plateaus thereafter. We report mean and standard error over 5 seeds.}
    \label{tab:num_comp_scaling}
\end{minipage}
\end{table}

\section{Analysis}

\subsection{Scaling of Number of Diffusion Components}

\begin{table*}[t]
\centering
\scriptsize
\setlength{\tabcolsep}{4pt}
\resizebox{0.99\textwidth}{!}{
\begin{tabular}{l|cccccccc|c}
\toprule
Policy & Close Box & Close Drawer & Close Fridge & Close Microwave & Toilet Seat Down & Take Umbrella  & Reach Target & Pick \& Lift & Avg. \\ 
     \toprule
DP   & 33.5 {\tiny ± 1.52} & 100.0 {\tiny ± 0.00} & 77.0 {\tiny ± 1.64} & 65.5 {\tiny ± 1.48} & 64.0 {\tiny ± 2.19} & 19.0 {\tiny ± 1.95} & \phantom{0}3.5 {\tiny ± 0.89} & 2.0 {\tiny ± 0.84} & 45.6 {\tiny ± 0.43} \\
\ourshort  & 72.0 {\tiny ± 1.30} & 100.0 {\tiny ± 0.00} & 75.5 {\tiny ± 1.10} & 93.0 {\tiny ± 1.92} & 27.5 {\tiny ± 2.12} & 78.0 {\tiny ± 2.17} & 61.0 {\tiny ± 1.52} & 4.5 {\tiny ± 2.39} & 63.9 {\tiny ± 0.23} \\
\ourshort{}$_{\text{top2}}$ & 63.5 {\tiny ± 1.14} & \phantom{0}95.5 {\tiny ± 1.64} & 69.0 {\tiny ± 1.00} & 90.5 {\tiny ± 0.71} & 22.5 {\tiny ± 2.70} & 30.5 {\tiny ± 2.70} & 41.5 {\tiny ± 1.30} & 1.0 {\tiny ± 0.55} & 51.8 {\tiny ± 0.27}      \\ 
    \bottomrule
\end{tabular}
}
\caption{\textbf{Partial reconstruction on RLBench.} \ourshort{}$_{\text{top2}}$ uses only top-2 components, achieving a 2$\times$ speedup in inference time with only 19\% relative performance drop. Results report the mean and standard error over 5 seeds, with 40 samples evaluated per seed.}
\label{tab:fdp_top2}
\end{table*}

\begin{table*}[t]
\centering
\scriptsize
\setlength{\tabcolsep}{3.5pt}
\resizebox{0.99\textwidth}{!}{
\begin{tabular}{l|cccc|cccccccc|c}
\toprule
 & \multicolumn{4}{c|}{\textbf{Multitask 4 Experts}}
 & \multicolumn{8}{c|}{\textbf{+ Expert Modules (5–12)}} &  \\
Policy
& \begin{tabular}{@{}c@{}}L1 PnP\\ soup\end{tabular}
& \begin{tabular}{@{}c@{}}L2 PnP\\ cheese\end{tabular}
& \begin{tabular}{@{}c@{}}K2 open\\ top drawer\end{tabular}
& \begin{tabular}{@{}c@{}}K3 PnP\\ pot\end{tabular}

& \begin{tabular}{@{}c@{}}L2 PnP\\ soup\end{tabular}
& \begin{tabular}{@{}c@{}}L5 PnP\\ mug\end{tabular}
& \begin{tabular}{@{}c@{}}L6 PnP\\ mug\end{tabular}
& \begin{tabular}{@{}c@{}}K3 turn\\ on stove\end{tabular}
& \begin{tabular}{@{}c@{}}K4 PnP\\ bowl\end{tabular}
& \begin{tabular}{@{}c@{}}K6 PnP\\ mug\end{tabular}
& \begin{tabular}{@{}c@{}}K8 PnP\\ pot\end{tabular}
& \begin{tabular}{@{}c@{}}S1 PnP\\ book\end{tabular}

& \begin{tabular}{@{}c@{}}Avg.\end{tabular} \\
\midrule
SDP 
    & 57.2 {\tiny ± 1.56} & 22.4 {\tiny ± 1.04} & 100.0 {\tiny ± 0.00} & 50.8 {\tiny ± 1.34} & 30.4 {\tiny ± 0.67} & 56.0 {\tiny ± 1.50} & 46.4 {\tiny ± 0.88} & \phantom{0}94.8 {\tiny ± 2.63} & 56.0 {\tiny ± 1.13} & 22.4 {\tiny ± 0.36} & 25.6 {\tiny ± 1.73} & 35.6 {\tiny ± 1.43} & 49.8 {\tiny ± 0.44} \\
MoDE 
    & 65.2 {\tiny ± 1.84} & 33.6 {\tiny ± 0.36} & \phantom{0}94.8 {\tiny ± 0.72} & 45.6 {\tiny ± 1.91} & 30.8 {\tiny ± 0.91} & 63.6 {\tiny ± 0.67} & 41.6 {\tiny ± 1.91} & 100.0 {\tiny ± 0.00} & 58.8 {\tiny ± 1.21} & 35.6 {\tiny ± 1.54} & 42.4 {\tiny ± 1.54} & 50.8 {\tiny ± 1.21} & 55.2 {\tiny ± 0.30} \\
\ourshort 
    & 83.2 {\tiny ± 0.91} & 34.4 {\tiny ± 1.54} & 100.0 {\tiny ± 0.00} & 60.8 {\tiny ± 2.30} & 40.0 {\tiny ± 0.57} & 89.6 {\tiny ± 0.88} & 56.8 {\tiny ± 1.75} & 100.0 {\tiny ± 0.00} & 82.0 {\tiny ± 1.26} & 40.8 {\tiny ± 2.50} & 64.8 {\tiny ± 1.21} & 55.6 {\tiny ± 1.54} & 67.3 {\tiny ± 0.39} \\
\bottomrule
\end{tabular}
}
\caption{\textbf{Continual adaptation on LIBERO.} We pretrain 4-expert \ourshort on first 4 tasks, and progressively add new experts for each additional adaptation task, ultimately reaching 12 experts. Results report the mean and standard error over 5 seeds, with 50 samples evaluated per seed. PnP stands for pick \& place.}
\label{tab:cont_adapt}
\vspace{-10pt}
\end{table*}

We study how the number of diffusion components in \ourshort affects multitask performance. Experiments are conducted on selected tasks from MetaWorld (\textit{door close}, \textit{drawer close}, \textit{disassemble}, \textit{window open}) and RLBench (\textit{toilet seat up}, \textit{open box}, \textit{open drawer}, \textit{take umbrella out}). As shown in Table~\ref{tab:num_comp_scaling}, increasing the number of components from 2 to 4 consistently improves performance, reflecting greater expressiveness and better sub-skill specialization. Beyond 4 components, performance saturates, suggesting diminishing returns.  

We view the number of components as a policy-level hyperparameter, analogous to the number of transformer layers or experts in other architectures. Intuitively, if we consider a latent space of observation–action pairs from all demonstrations, the ideal number of components would correspond to the number of clusters in this space, with each component modeling a coherent subset of behaviors~\cite{liu2023unsupervisedcompositionalconceptsdiscovery}. In practice, we find that 4–6 components provide a good trade-off between model complexity and performance for the tasks considered. While inference cost scales linearly with the number of components, \ourshort supports deployment optimizations such as pruning (see Section~\ref{sec:distr_partial_reconst}), distillation, or merging~\cite{biggs2024diffusionsoupmodelmerging} to reduce overhead.

\subsection{Scaling of Number of Tasks} 

We further evaluate how the relative advantage of \ourshort scales with the number of tasks. Fig.~\ref{fig:num_task_scaling} shows the relative success rate improvement of \ourshort over DP as the multitask setting becomes more challenging. The performance gap widens with more tasks, highlighting that \ourshort~’s modular factorization is particularly effective in modeling increasingly complex and multimodal action distributions.

\subsection{Scaling of Number of Demonstrations}

\begin{figure}[t]
    \centering
    \begin{subfigure}{.85\columnwidth}
        \caption{Multitask Learning}
        \includegraphics[width=\columnwidth]{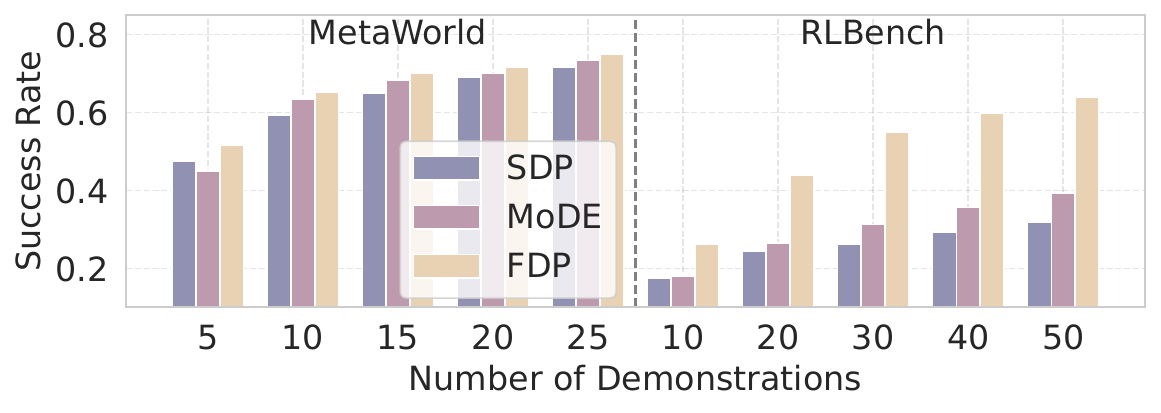}
        \label{fig:num_demo_multitask_scaling}
        \vspace{-15pt}
    \end{subfigure}
    \vfill
    \begin{subfigure}{.85\columnwidth}
        \caption{Task Adaptation}
        \includegraphics[width=\columnwidth]{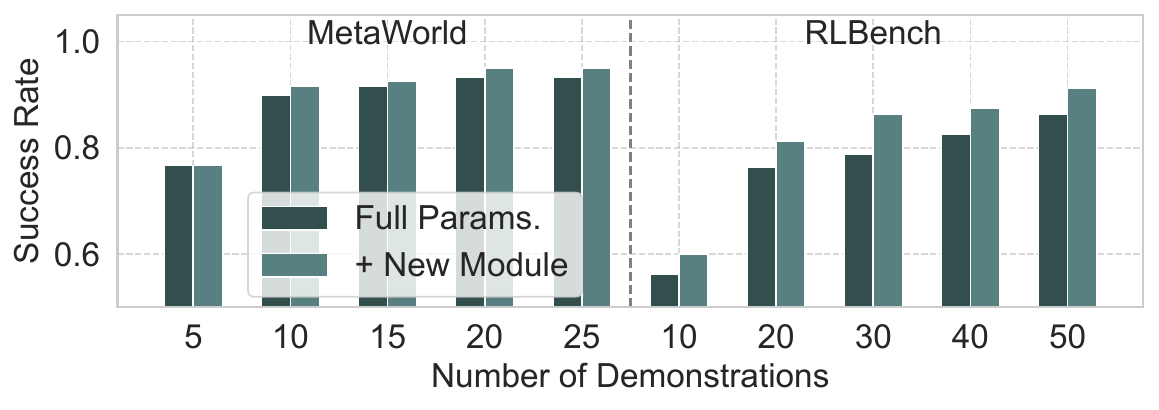}
        \label{fig:num_demo_adaptation_scaling}
        \vspace{-15pt}
    \end{subfigure}
    \caption{\textbf{Performance scaling with number of demonstrations.} \textbf{(a)} Metaworld tasks \textit{door open, drawer open, assembly, window close, peg insert, hammer}; RLBench tasks \textit{door open, drawer open, assembly, window close, peg insert, hammer}. \textbf{(b)} Metaworld tasks \textit{door close, drawer close, disassemble, window open}; RLBench tasks \textit{toilet seat down, close box}}
\end{figure}

\subsubsection{\textbf{Multitask Learning}}
We evaluate how \ourshort benefits from increasing amounts of demonstration data. As shown in Fig.~\ref{fig:num_demo_multitask_scaling}, performance improves steadily with more demonstrations. \ourshort consistently outperforms baselines, with particularly large gains on RLBench where complex, contact-rich interactions make effective decomposition especially valuable.

\subsubsection{\textbf{Task Adaptation}}
We analyze how adaptation performance scales with the number of demonstrations, and compare the proposed \textit{+ New Module} strategy with full-parameter fine-tuning. As shown in Fig.~\ref{fig:num_demo_adaptation_scaling}, both strategies benefit from more data, but \textit{+ New Module} achieves comparable or better performance with even fewer demonstrations (on RLBench). This highlights the strength of our modular design in enabling data-efficient adaptation while avoiding the cost and potential catastrophic forgetting associated with updating all model parameters.

\subsection{Partial Reconstruction of Action Distribution}
\label{sec:distr_partial_reconst}

\ourshort supports MoE-style pruning by composing only the top-$k$ components at inference, effectively performing a \textit{partial reconstruction} of the action distribution. Instead of aggregating all components, we sample from the distribution reconstructed by the most relevant ones, preserving the modes most critical for the current observation. This focuses computation on the most informative components, reducing cost while maintaining strong performance. As shown in Table~\ref{tab:fdp_top2}, using only the top-2 components (\ourshort{}$_\text{top2}$) results in a 17\% relative performance drop but substantially improves inference speed, requiring no retraining and only a minor change to the sampling code.

\subsection{Continual Adaptation}

To evaluate scalability of adaptation, we construct a continual adaptation benchmark with 12 LIBERO tasks. Starting from a 4-expert \ourshort pretrained on 4 tasks, we sequentially introduce one new expert per new task, freezing previous modules throughout. This setup results in a 12-expert policy by the end. Table~\ref{tab:cont_adapt} demonstrates that \ourshort consistently outperforms SDP and MoDE across all stages, maintaining high success rates as additional modules are added. We observe that despite the growing number of frozen components, both training and inference remain stable. The router effectively identifies and leverages relevant experts, even in the presence of many potentially redundant or unused experts. This result demonstrates that \ourshort enables fast, scalable, modular adaptation.

\subsection{Knowledge Retention}

\begin{table}[t]
\centering
\scriptsize
\setlength{\tabcolsep}{3.5pt}
\begin{tabular}{l|cccc|c}
\toprule
Policy
& \begin{tabular}{@{}c@{}}L1 PnP\\ soup\end{tabular}
& \begin{tabular}{@{}c@{}}L2 PnP\\ cheese\end{tabular}
& \begin{tabular}{@{}c@{}}K2 open\\ top drawer\end{tabular}
& \begin{tabular}{@{}c@{}}K3 PnP\\ pot\end{tabular}

& \begin{tabular}{@{}c@{}}Avg.\end{tabular} \\
\midrule
\ourshort{}$_{\text{pretrain}}$
    & 57.2 {\tiny ± 1.56} & 22.4 {\tiny ± 1.04} & 100.0 {\tiny ± 0.00} & 50.8 {\tiny ± 1.34} & 57.6 {\tiny ± 0.61} \\
\ourshort{}$_{\text{w/o buffer}}$
    & 32.5 {\tiny ± 1.00} & \phantom{0}5.0 {\tiny ± 1.22} & \phantom{0}50.5 {\tiny ± 1.10} & 29.0 {\tiny ± 1.14} & 29.3 {\tiny ± 0.45} \\
\ourshort{}$_{\text{w/ buffer}}$
    & 53.0 {\tiny ± 1.64} & 17.5 {\tiny ± 1.00} & \phantom{0}96.0 {\tiny ± 0.89} & 47.5 {\tiny ± 1.22} & 53.5 {\tiny ± 0.63} \\
\bottomrule
\end{tabular}
\caption{\textbf{Knowledge Retention Analysis on LIBERO.} We report success rates on four pretraining tasks after adapting policies to two new tasks (\textit{L2 PnP soup} and \textit{L5 PnP mug}). We compare: (i) \ourshort{}$_{\text{pretrain}}$: the original model before adaptation; (ii) \ourshort{}$_{\text{w/o buffer}}$: the model after adaptation using only new task data; and (iii) \ourshort{}$_{\text{w/ buffer}}$: the model after adaptation using a small replay buffer (5 demos/task) to mitigate forgetting. Results show mean and standard error over 5 seeds, 50 evaluations per seed. PnP stands for pick \& place.}
\label{tab:anti_forget}
\end{table}

We investigate the model's ability to retain knowledge from base tasks during adaptation, a critical property for scalable lifelong learning systems. Conceptually, since the pretrained diffusion experts in \ourshort remain frozen, the core motor skills are inherently preserved. While a distribution shift in the observation encoder (which is trained from scratch in this work) or a reallocation of weights by the router can occur, these effects can be mitigated by employing a frozen vision foundation model like CLIP~\cite{radford2021clip} or by caching the minimal router checkpoints. More generally, to support \ourshort as a lifelong learning system, we evaluate a strategy using a small replay buffer containing 5 demonstrations per pretraining task. As shown in Table~\ref{tab:anti_forget}, while adaptation without a buffer leads to some performance degradation, the inclusion of a minimal buffer allows the model to retain nearly all of its original performance. These results suggest that factorization provides a robust foundation for knowledge retention, with further investigation into long-term lifelong learning dynamics reserved for future work.

\subsection{Diffusion Components Analysis}
\label{sec:diff_comp_analysis}

\addtocounter{figure}{-1}   
\begin{figure}[t]
    \centering

    \begin{subfigure}{0.49\linewidth}
        \centering
        \caption{Assembly}
        \includegraphics[width=\linewidth]{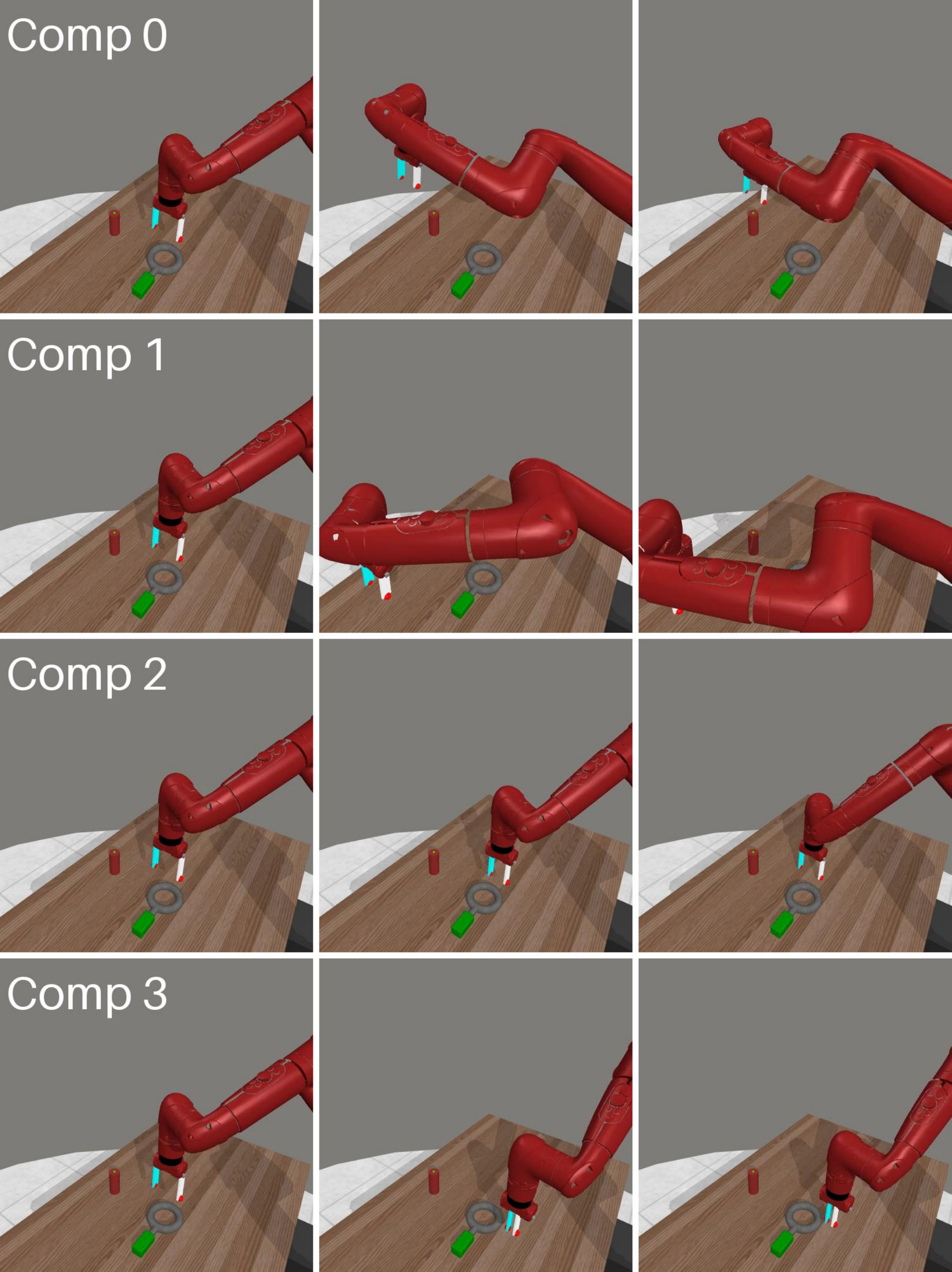} \\
        \vspace{5pt}
        \includegraphics[width=\linewidth]{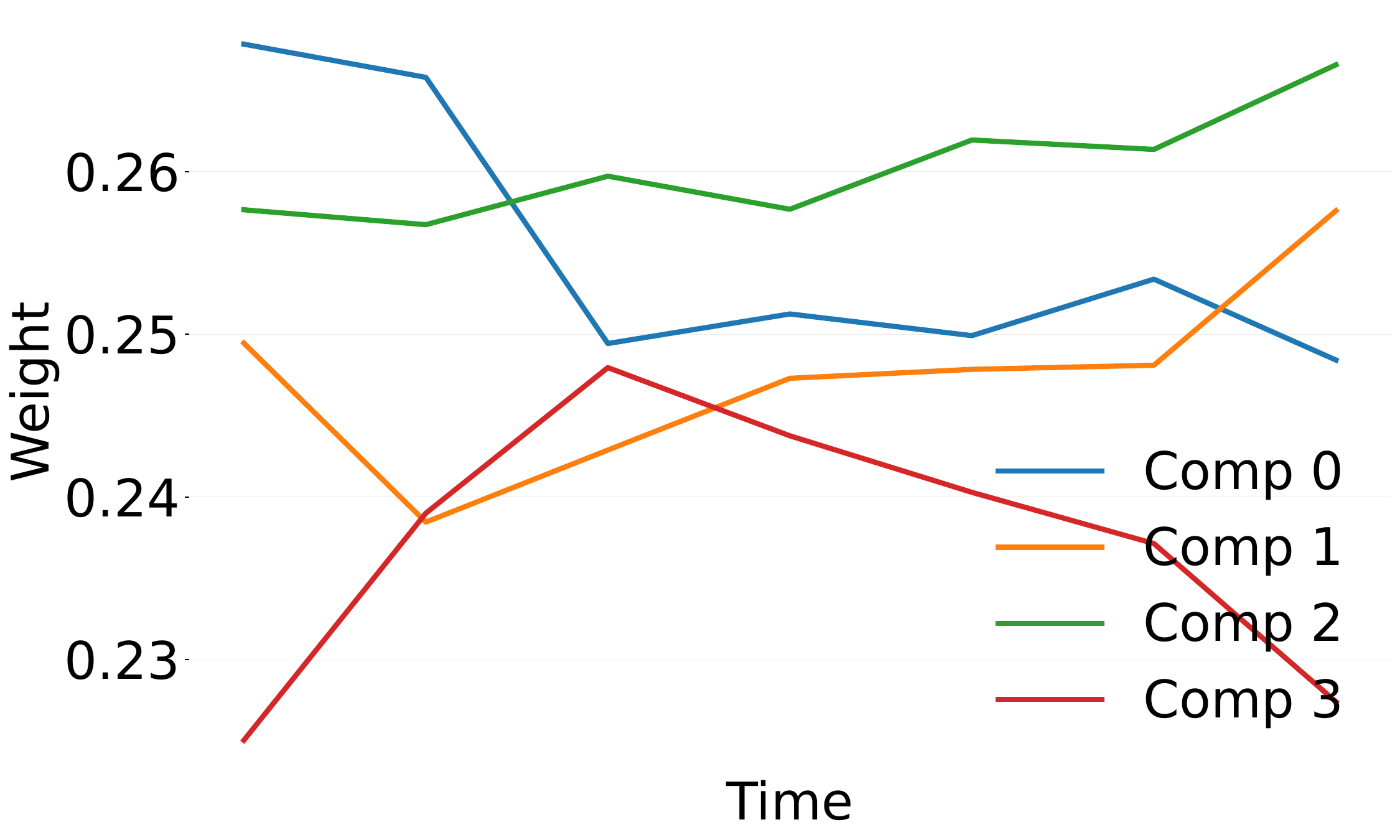}
    \end{subfigure}
    \hfill
    \begin{subfigure}{0.49\linewidth}
        \centering
        \caption{Hammer}
        \includegraphics[width=\linewidth]{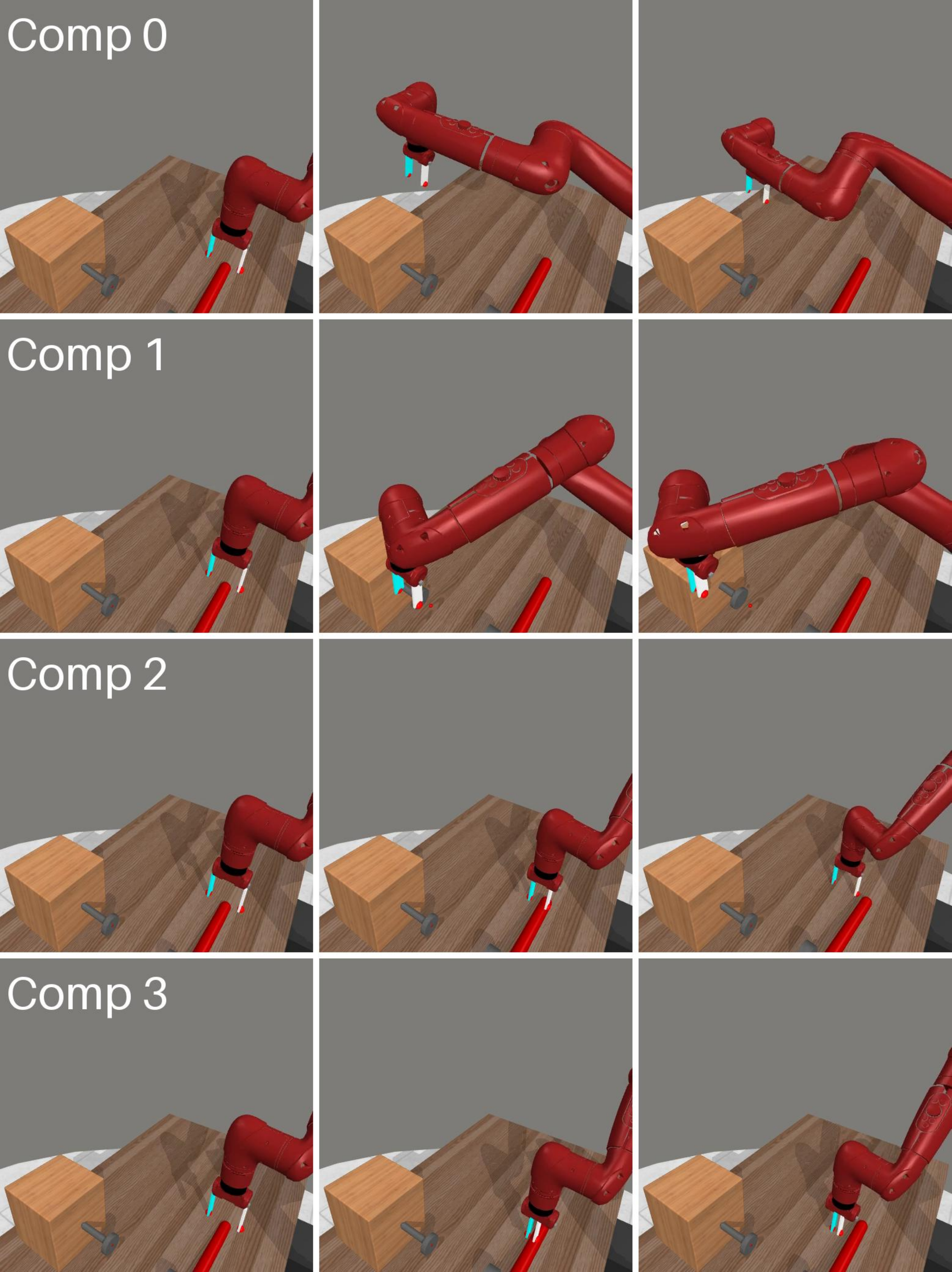} \\
        \vspace{5pt}
        \includegraphics[width=\linewidth]{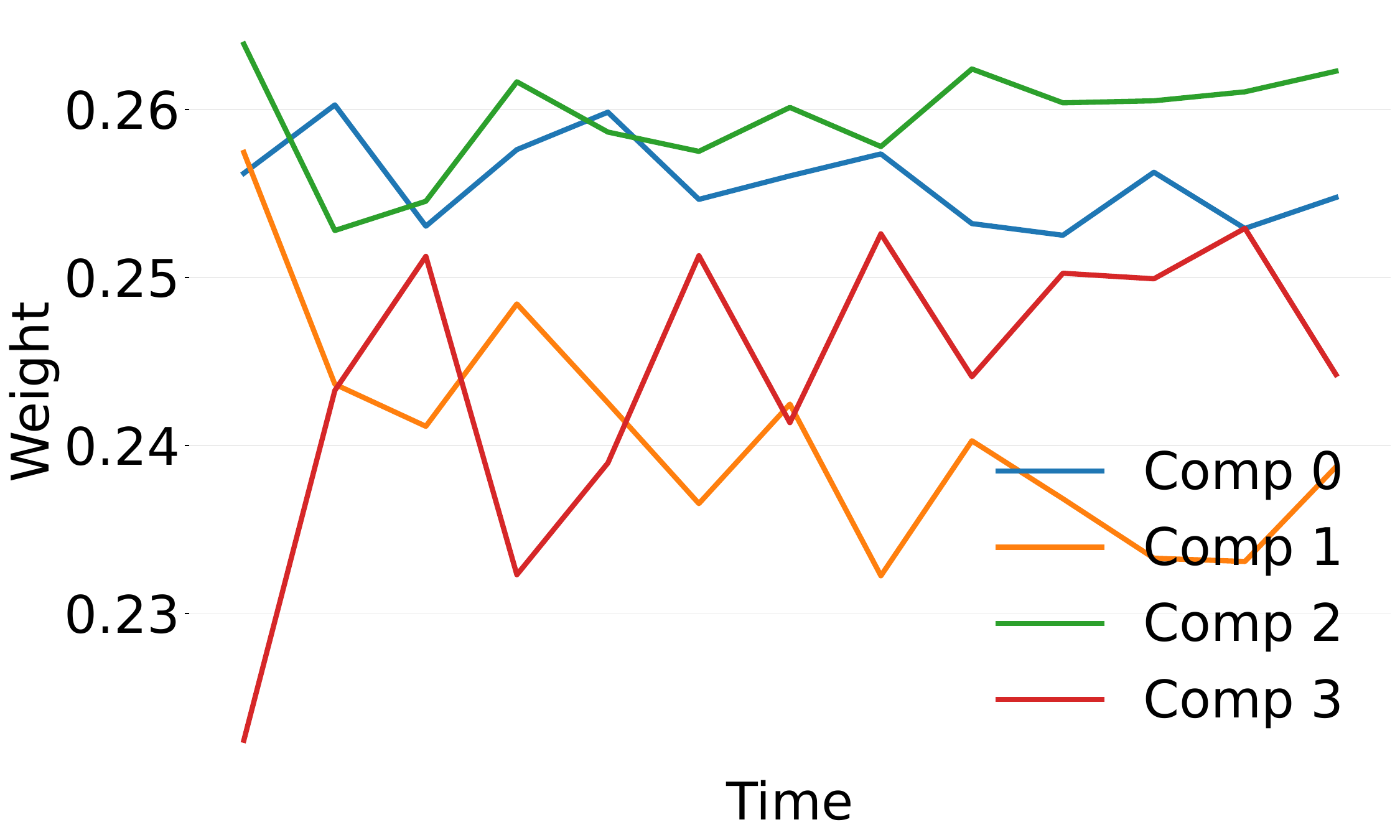}
    \end{subfigure}
    
    \vspace{-5pt}
    \captionof{figure}{\textbf{Rollout trajectories of individual diffusion components} in \ourshort. 
    \textbf{(a)} In \textit{assembly}, components 0 and 1 align the robot with the stand, component 2 aligns with the ring, and component 3 executes the grasp. 
    \textbf{(b)} In \textit{hammer}, components 0 and 1 align and approach the pin, component 2 approaches the hammer, and component 3 performs the grasp.}
    \label{fig:vis_comp_rollout}
\end{figure}

\begin{figure}[t]
    \centering
    \includegraphics[width=.8\linewidth]{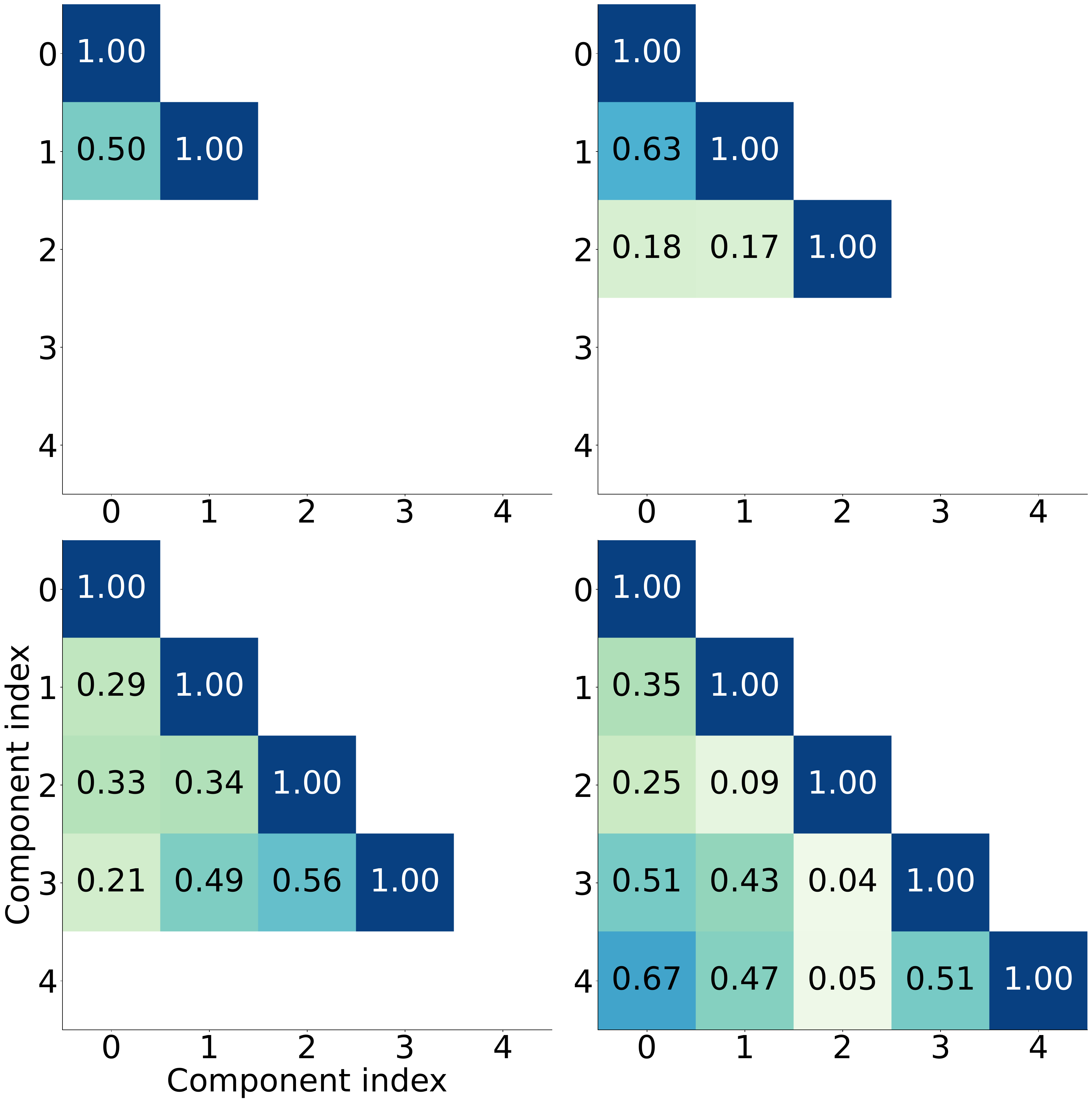}
    \caption{\textbf{Cosine similarity between diffusion component scores.} Each heatmap visualizes average pairwise similarity for independent \ourshort instances configured with 2-5 components, computed over four RLBench tasks. Lower similarity indicates more distinct behavioral specialization. Note that subplots represent separate training runs rather than a single evolving model.}
    \label{fig:score_cossim}
    \vspace{-15pt}
\end{figure}

\begin{figure}[t]
    \centering
    \includegraphics[width=.95\columnwidth]{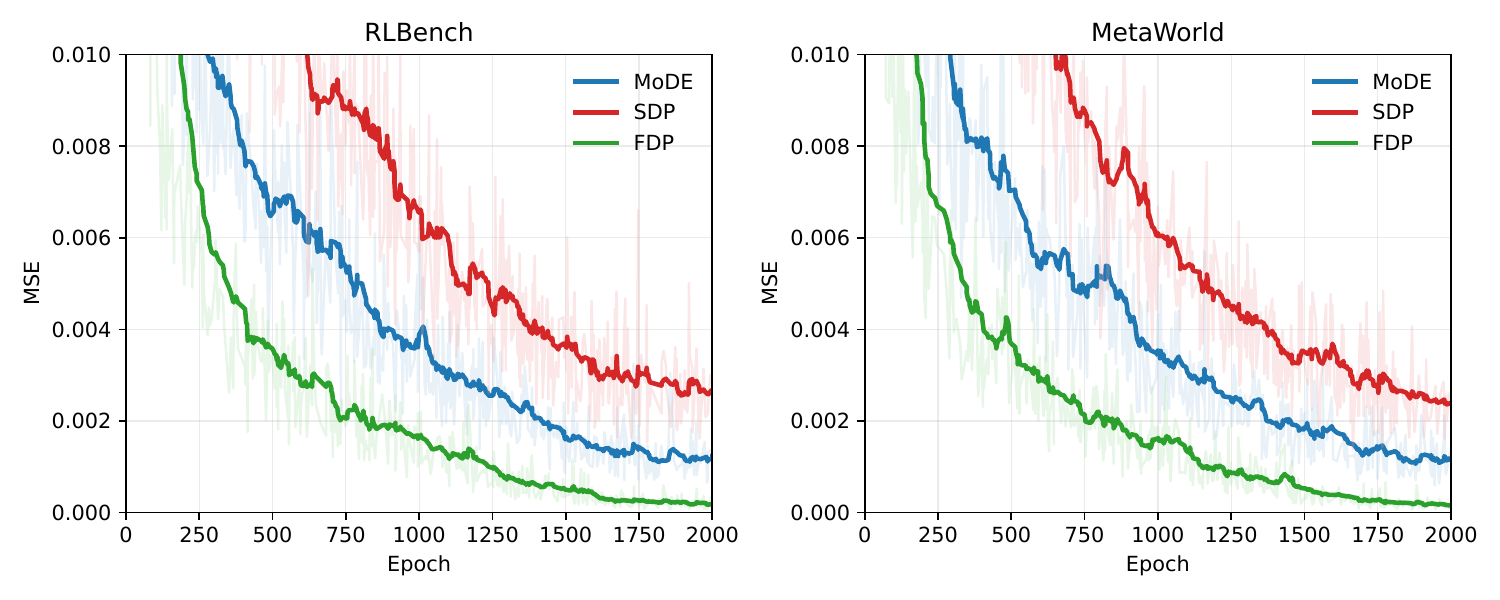}
    \vspace{-7pt}
    \caption{\textbf{Training convergence curves}. Mean squared error (MSE) loss over training epochs for RLBench and MetaWorld tasks. \ourshort consistently converges faster and more stably than MoDE and SDP, indicating improved training efficiency and optimization stability.}
    \label{fig:training_convergence}
\end{figure}

To better understand how modularity manifests in \ourshort, we analyze the behavior and specialization of individual diffusion components. Fig.~\ref{fig:vis_comp_rollout} shows rollout trajectories and activation weights produced by each component in two representative MetaWorld tasks: \textit{assembly} and \textit{hammer}. Across both tasks, we observe that different components specialize in distinct functional stages, such as alignment, approach, and grasp execution. Notably, the weights for Component 3 (responsible for gripper closure) align with task phases: in \textit{assembly}, the weight increases as the robot grasps the ring and decreases after placement;  in \textit{hammer}, the weight increases during the initial grasp and remains elevated as the robot must consistently hold the hammer to strike the pin. This suggests that \ourshort naturally decomposes complex behaviors into distinct sub-skills across its components.

To complement the qualitative analysis, we compute the pairwise cosine similarity between the score outputs, shown in Fig.~\ref{fig:score_cossim}, visualize how the learned components relate to each other during inference. While the components are not completely orthogonal, we observe noticeable variation between different pairs, indicating that diffusion components capture distinct, though partially overlapping, aspects of the behavior distribution.

\ourshort's structure contrasts with baseline MoE-based policies. In MoDE, experts specialize according to diffusion noise levels rather than task semantics, leading to noise-level specialization that lacks behavioral interpretability. In SDP, sub-skills emerge from sets of experts selected across layers, making it difficult to assign functionality to any single expert. Experts can be reused across different combinations or ignored altogether. Furthermore, SDP routers tend to favor a small subset of experts, leading to poor load balancing and limited diversity~\cite{wang2024sparse_dp_moe}. \ourshort assigns each behavioral mode to a distinct, standalone diffusion component. This avoids routing instability or expert redundancy commonly seen in MoE. This modular design facilitates straightforward analysis and reuse, contributing to better training stability and more coherent specialization.

\subsection{Training Convergence of Policies}

We compare the training efficiency of \ourshort against MoDE and SDP by analyzing convergence curves on validation trajectories. Specifically, we track the mean squared error (MSE) loss used during diffusion training, measured over validation episodes across training epochs. Results are shown in Fig.~\ref{fig:training_convergence} for both MetaWorld and RLBench tasks.

\ourshort consistently achieves lower validation MSE in fewer epochs, indicating faster convergence. MoDE converges more slowly, while SDP shows higher variance and slower reduction in loss, likely due to instability in expert selection and poor load balancing during training. These results support our claim that continuous score composition in \ourshort improves optimization stability compared to discrete MoE methods.


\section{Conclusion And Limitations}

We present \ourshort, a modular policy architecture that leverages factorized diffusion models for multitask imitation learning and efficient task adaptation. By composing behavior-specialized diffusion components, our method improves generalization, interpretability, and modularity over prior approaches. Extensive experiments on both simulated and real-world tasks demonstrate that \ourshort outperforms strong baselines in multitask performance and adapts effectively to new tasks.

While our work demonstrates clear modular specialization, there remain interesting directions for further analysis. First, we currently use homogeneous diffusion components of similar architecture and size. Future work could explore heterogeneous module designs, such as mixing U-Net and Transformer-based diffusion models, or using modules of varying sizes, to enhance flexibility and expressiveness. Second, we primarily study specialization through rollout visualization; an alternative approach is to systematically remove individual diffusion components and observe the resulting policy behaviors and failure modes. This could provide deeper insights into the roles and dependencies of different sub-skills captured by the factorized policy.


\bibliographystyle{plainnat}
\bibliography{reference}

\end{document}